\documentclass[preprint,5p,12pt,twocolumn]{article}

\usepackage{amssymb}
\usepackage{subfig}
\usepackage{algorithm,algorithmic}
\usepackage{graphicx}
\usepackage{caption}
\usepackage{xspace}
\usepackage{multirow}
\usepackage{rotating}
 \usepackage{url}
\usepackage{gensymb}
\usepackage [ table ]{ xcolor }
\usepackage[top=2cm, bottom=2cm, left=1.5cm, right=1.5cm]{geometry}
\usepackage{setspace}
\usepackage{morefloats}

\usepackage[affil-it]{authblk}

\begin{document}

\title{Deep Networks Can Resemble  Human Feed-forward Vision in Invariant Object Recognition}

\author{Saeed Reza Kheradpisheh$ ^{1,6} $ }
\author{Masoud Ghodrati$ ^{2} $ }
\author{Mohammad Ganjtabesh$ ^{1,} $\thanks{Corresponding author.\\ Email addresses:\\ kheradpisheh@ut.ac.ir (SRK),\\masoud.ghodrati@monash.edu (MGh) \\ mgtabesh@ut.ac.ir (MG),\\ timothee.masquelier@alum.mit.edu (TM).}}
\author{Timoth\'ee Masquelier$ ^{3,4,5,6,*} $}

\affil{\footnotesize $ ^{1} $ Department of Computer Science, School of Mathematics, Statistics, and Computer Science, University of Tehran, Tehran, Iran
\\ $ ^{2} $ Department of Physiology, Monash University, Melbourne, VIC, Australia\\ $ ^{3} $ INSERM, U968, Paris, F-75012, France\\ $ ^{4} $ Sorbonne Universit\'es, UPMC Univ Paris 06, UMR-S 968, Institut de la Vision, Paris, F-75012, France\\ $ ^{5} $ CNRS, UMR-7210, Paris, F-75012, France\\ $ ^{6} $ CERCO UMR 5549, CNRS – Universit\'e de Toulouse, F-31300, France}
\date{}

\maketitle
\begin{abstract}
Deep convolutional neural networks (DCNNs) have attracted much attention recently, and have shown to be able to recognize thousands of object categories in natural image databases. Their architecture is somewhat similar to that of the human visual system: both use restricted receptive fields, and a hierarchy of layers which progressively extract more and more abstracted features. Yet it is unknown whether DCNNs match human performance at the task of view-invariant object recognition, whether they make similar errors and use similar representations for this task, and whether the answers depend on the magnitude of the viewpoint variations. To investigate these issues, we benchmarked eight state-of-the-art DCNNs, the HMAX model, and a baseline shallow model and compared their results to those of humans with backward masking. Unlike in all previous DCNN studies, we carefully controlled the magnitude of the viewpoint variations to demonstrate that shallow nets can outperform deep nets and humans when variations are weak. When facing larger variations, however, more layers were needed to match human performance and error distributions, and to have representations that are consistent with human behavior. A very deep net with 18 layers even outperformed humans at the highest variation level, using the most human-like representations.


\end{abstract}
\section*{Introduction}

Primates excel at view-invariant object recognition~\cite{dicarlo2012does}. This is a computationally demanding task, as an individual object can lead to an infinite number of very different projections onto the retinal photoreceptors while it varies under different 2-D and 3-D transformations. It is believed that the primate visual system solves the task through hierarchical processing along the ventral stream of the visual cortex~\cite{dicarlo2012does}. This stream ends in the inferotemporal cortex (IT), where object representations are robust, invariant, and linearly-separable~\cite{dicarlo2007untangling,dicarlo2012does}. Although there are extensive within- and between-area feedback connections in the visual system, neurophysiological~\cite{liu2009timing,freiwald2010functional}, behavioral~\cite{thorpe1996speed}, and computational~\cite{anselmi2014unsupervised} studies suggest that the first feed-forward flow of information ($\sim 100-150$ ms post-stimulus presentation) might be sufficient for object recognition~\cite{thorpe1996speed,hung2005fast} and even invariant object recognition~\cite{liu2009timing,freiwald2010functional,anselmi2014unsupervised,hung2005fast}.

Motivated by this feed-forward information flow and the hierarchical organization of the visual cortical areas, many computational models have been developed over the last decades to mimic the performance of the primate ventral visual pathway in object recognition. Early models were only comprised of a few layers~\cite{Fukushima1980,LeCun1998,Serre2007.PAMI,Masquelier2007,Lee2009}, while the new generation, called ``deep convolutional neural networks" (DCNNs) contain many layers (8 and above). DCNNs are large neural networks with millions of free parameters that are optimized through an extensive training phase using millions of labeled images~\cite{cox2014neural}. They have shown impressive performances in difficult object and scene categorization tasks with hundreds of categories~\cite{schmidhuber2015deep,cox2014neural,Krizhevsky2012,zeiler2014visualizing,sermanet2013overfeat,chatfield2014return}. Yet the view-point variations were not carefully controlled in these studies. This is an important limitation: in the past, it has been shown that models performing well on apparently challenging image databases may fail to reach human-level performance when objects are varied in size, position, and most importantly 3-D transformations~\cite{ghodrati2014feedforward,khaligh2014deep,pinto2011comparing,Pinto2008}. DCNNs are  position invariant by construction, thanks to weight sharing. However, for other transformations such as scale, rotation in depth, rotation in plane, and 3-D transformations, there is no built-in invariance mechanism. Instead, these invariances are acquired through learning. Although the features extracted by DCNNs are significantly more powerful than their hand-designed counterparts like SIFT and HOG~\cite{khaligh2014deep,Liu2015}, they may have difficulties to tackle  3-D transformations.

To date, only a handful of  studies have assessed the performance of DCNNs and their constituent layers in invariant object recognition \cite{yosinski2014transferable,peng2014exploring,cheung2014discovering,khaligh2014deep,cadieu2014deep,gucclu2014deep}. In this study we systematically compared humans and DCNNs at view-invariant object recognition, using exactly the same images. The advantages of our work with respect to  previous studies are: (1) we used a larger object database, divided into five categories; (2) most importantly,  we controlled and varied the magnitude of the variations in size, position, in-depth and in-plane rotations; (3) we benchmarked eight state-of-the-art DCNNs, the HMAX model~\cite{Serre2007.PAMI} (an early biologically inspired  shallow model), and a very simple shallow model that classifies directly from the pixel values ("Pixel"); (4) in our psychophysical experiments, the images were presented briefly and with backward masking, presumably blocking feedback; (5) we performed extensive comparisons between different layers of DCNNs and studied how invariance evolves through the layers; (6) we compared models and humans in terms of performance, error distributions, and representational geometry; and (7) to measure the influence of the background on the invariant object recognition problem our dataset included both segmented and unsegmented images.

This approach led to new findings: (1) Deeper was usually better and more human-like, but only in the presence of large variations; (2) Some DCNNs reached human performance even with large variations; (3) Some DCNNs had error distributions which were indiscernible from those of humans; (4) Some DCNNs used representations that were more consistent with human responses, and these were not necessarily the top performers.

\section*{Materials and methods}
\subsection*{Deep convolutional neural networks (DCNNs)}
The idea behind DCNNs is a combination of deep learning~\cite{schmidhuber2015deep} with convolutional neural networks~\cite{LeCun1998}. DCNNs have a hierarchy of several consecutive feature detector layers. Lower layers are mainly selective to simple features while higher layers tend to detect more complex features. Convolution is the main process in each layer that is generally followed by complementary operations such as max pooling and output normalization. Up to now, various learning algorithms have been proposed for DCNNs, and among them the supervised learning methods have achieved stunning successes\cite{lecun2015deep}. Recent advances have led to the birth of supervised DCNNs with remarkable performances on extensively large and difficult object databases such as Imagenet~\cite{lecun2015deep,schmidhuber2015deep}. We have selected  the eight most recent, powerful, and supervised DCNNs and tested them in one of the most challenging visual recognition task, i.e. invariant object recognition. Below are short descriptions of all the DCNNs that we studied in this work.

\paragraph{Krizhevsky et. al. 2012} This outstanding model reached an impressive performance on the Imagenet database and significantly defeated other competitors in the ILSVRC-2012 competition~\cite{Krizhevsky2012}. The excellent performance of this model attracted  attention towards the abilities of DCNNs and opened a new avenue for further investigations. Briefly, the model contains five convolutional (feature detector) and three fully connected (classification) layers. They used the Rectified Linear Units (ReLUs) for the neurons' activation function, which significantly speeds up the learning phase. The max pooling operation is performed in the first, second, and fifth convolutional layers. This model is trained using a stochastic gradient descent algorithm. It has about 60 million free parameters; to avoid overfitting, they used some data augmentation techniques to enlarge the training set as well as the dropout technique in the learning procedure of the first two fully-connected layers. The structural details of this model are presented in Table~\ref{table1}. We used the pre-trained version of this model (on the Imagenet database) which is publicly released at \url{http://caffe.berkeleyvision.org} by Jia et.~al~\cite{jia2014caffe}.

\paragraph{Zeiler and Fergus 2013} To better understand the ongoing functions of different layers in Krizhevsky's model, Zeiler and Fergus~\cite{zeiler2014visualizing} introduced a deconvolutional visualizing technique which reconstructs the  features learned by each neuron. This enabled them to detect and resolve deficiencies by optimizing  architecture and parameters of the Krizhevsky model. Briefly, the visualization showed that the neurons of the first two layers were mostly converged to extremely high and low frequency information. Besides, they detected aliasing artifacts caused by the large stride in the second convolutional layer. To resolve these issues, they reduced the first layer filter size, from $11 \times 11$ to $7\times7$, and decreased the stride of the convolution in the second layer from 4 to 2. The results showed a reasonable performance improvement with respect to the Krizhevsky model. The structural details of this model are provided in Table~\ref{table1}. We used the Imagenet pre-trained version of Zeiler and Fergus model available at \url{http://libccv.org}.

\paragraph{Overfeat 2014} The Overfeat model~\cite{sermanet2013overfeat} provides a complete system to do  object classification and  localization together. Overfeat has been proposed in two different types: the \textit{Fast} model with eight layers and the \textit{Accurate} model with nine layers. Although the number of free parameters in both types are nearly the same (about 145 million), there are about twice as many connections in the \textit{Accurate} one. It has been shown that the \textit{Accurate} model leads to a better performance on Imagenet than the \textit{Fast} one. Moreover, after the training phase, to make  decisions with optimal confidence and increase the final accuracy, the classification can be performed in different scales and positions. Overfeat has some important differences with other DCNNs: 1) there is no local response normalization, 2) the pooling regions are non-overlapping, and 3) the model has smaller convolution stride ($=2$) in  the first two layers. The specifications of the \textit{Accurate} version of the Overfeat model, which we used in this study, are presented in Table~\ref{table1}.  Similarly, we used the Imagenet pre-trained model which is publicly available at \url{http://cilvr.nyu.edu/doku.php?id=software:Overfeat:start}.

\paragraph{Hybrid-CNN 2014} The Hybrid-CNN model~\cite{zhou2014learning} has been designed to do a scene-understanding task. This model was trained on ~3.6 million images of 1183 categories including 205 scene categories from the place database and 978 object categories from the training data of the Imagenet database. The scene labeling, which consists of some fixed descriptions about the scene appearing in each image, was performed by a huge number of Amazon Mechanical Turk workers.  The overall structure of Hybrid-CNN is similar to the Krizhevsky model (see Table~\ref{table1}), but it is trained on a different dataset to perform a scene understanding task. This model is publicly released  at  \url{http://places.csail.mit.edu}. Surprisingly, the hybrid-CNN significantly outperforms the Krizhevsky model in different scene-understanding benchmarks, while they perform similarly different object recognition benchmarks. 

\paragraph{Chatfield CNNs} Chatfield et.~al.~\cite{chatfield2014return} did an extensive comparison among the shallow and deep image representations. To this end, they proposed three different DCNNs with different architectural characteristics,  each exploring a different accuracy/speed trade-off. All three models have five convolutional and three fully connected layers but with different structures. The \textit{Fast} model (CNN-F) has smaller convolutional layers and the convolution stride in the first layer is four, versus 2 for CNN-M and -S, which leads to  a higher processing speed in the CNN-F model. The stride and receptive field of the first convolutional layer is decreased in \textit{Medium} model (CNN-M), which was shown to be effective for the Imagenet database\cite{zeiler2014visualizing}. The CNN-M model  also  has a larger stride in the second convolutional layer to reduce the computation time. The \textit{Slow} model (CNN-S) uses $ 7 \times 7 $ filters with stride of 2 in the first layer and larger max pooling window in the third and fifth convolutional layers. All these models were trained over the Imagenet database using a gradient descent learning algorithm. The training phase was performed over random crops sampled from the whole parts of the image rather than the central region. Based on the reported results, the performance of CNN-F model was close to the Zeiler and Fergus model while both CNN-M and CNN-S outperformed the Zeiler and Fergus model. The structural details of these three models are also presented in Table~\ref{table1}. All these models are available at \url{http://www.robots.ox.ac.uk/~vgg/software/deep_eval}.
\paragraph{Very Deep 2014}
Another important aspect of DCNNs is the number of internal layers, which influences their final performance. Simonyan and Zisserman~\cite{simonyan2014very} have studied the impacts of the network depth  by implementing  deep convolutional networks with 11, 13, 16, and 19 layers. To this end, they used very
small ($3 \times 3$) convolution filters in all layers, and steadily increased the depth of the network by adding more convolutional layers. Their results indicate that the recognition accuracy increases by adding more layers and the 19-layer model significantly outperformed other DCNNs. They have shown that their 19-layered model, trained on the Imagenet database, achieved high performances on other datasets without any fine-tuning. Here we used the 19-layered model available at \url{http://www.robots.ox.ac.uk/~vgg/research/very_deep/}. The structural details of this model are provided in Table~\ref{table1}.
\begin{table*}[t]

\caption{{\bf The architecture and settings  of different layers of DCNN models.} Each row of the table refers to a DCNN model and each column contains the details of a layer. The details of convolutional layers (labeled as Conv) are given in three sub-rows: the first one indicates the number and the size of the convolution filters as $ Num \times Size \times Size $; the convolution stride is given in the second sub-row; and the third one indicates the max pooling down-sampling rate, and if Linear Response Normalization (LRN) is used. The details of fully connected layers (labeled as Full) are presented in two sub-rows: the first one indicates the number of neurons; and the second one whether dropout or soft-max operations are applied.}
\label{table1}
\centering
\tiny
\resizebox{\textwidth}{!} {
\begin{tabular}{|c|c|c|c|c|c|c|c|c|c|c|}
\hline
Model& Layer 1&Layer 2&Layer 3&Layer 4&Layer 5&Layer 6&Layer 7&Layer 8&Layer 9&Layer 10\\
\hline
 & Conv&Conv&Conv&Conv&Conv&‌Full&‌Full&‌Full&&\\

Krizhevsky &$96\times 11\times 11$& $256\times 5\times 5$&$384\times 3\times 3$&$384\times 3\times 3$&$256\times 3\times 3$&4096&4096&1000&-&-\\
et.~al. 2012 & Stride 4& Stride 1& Stride 1&Stride 1&Stride 1&drop out&drop out& soft max&&\\
&LRN, x3 Pool&LRN, x3 Pool& -&-&x3 Pool&&& &&\\

\hline
 & Conv&Conv&Conv&Conv&Conv&‌Full&‌Full&‌Full&&\\

Zeiler and &$96\times 7\times 7$& $256\times 5\times 5$&$384\times 3\times 3$&$384\times 3\times 3$&$256\times 3\times 3$&4096&4096&1000&-&-\\
 Fergus 2013 & Stride 2& Stride 2& Stride 1&Stride 1&Stride 1&drop out&drop out& soft max&&\\
 & LRN, x3 Pool&LRN, x3 Pool& -&-&x3 Pool&&& &&\\
 
\hline
 & Conv&Conv&Conv&Conv&Conv&conv&‌Full&‌Full&‌Full&\\

OverFeat &$96\times 7\times 7$& $256\times 7\times 7$&$512\times 3\times 3$&$512\times 3\times 3$&$1024\times 3\times 3$&$1024\times 3\times 3$&4096&4096&1000&-\\
2014 & Stride 2& Stride 1& Stride 1&Stride 1&Stride 1&Stride 1&drop out&drop out &soft max&\\
 &  x3 Pool& x2 Pool& -&-&-&x3 Pool&& &&\\
 
\hline
 & Conv&Conv&Conv&Conv&Conv&‌Full&‌Full&‌Full&&\\

Hybrid-CNN &$96\times 11\times 11$& $256\times 5\times 5$&$384\times 3\times 3$&$384\times 3\times 3$&$256\times 3\times 3$&4096&4096&1183&-&-\\
2014 & Stride 4& Stride 1& Stride 1&Stride 1&Stride 1&drop out&drop out& soft max&&\\
&LRN, x3 Pool&LRN, x3 Pool& -&-&x3 Pool&&& &&\\

\hline
 & Conv&Conv&Conv&Conv&Conv&‌Full&‌Full&‌Full&&\\

 CNN-F&$64\times 11\times 11$& $256\times 5\times 5$&$256\times 3\times 3$&$256\times 3\times 3$&$256\times 3\times 3$&4096&4096&1000&-&-\\
2014&  Stride 4& Stride 1&Stride 1&Stride 1&Stride 1&drop out&drop out &soft max&&\\
&LRN, x2 Pool&LRN, x2 Pool& -&-&x2 Pool&&&&&\\
\hline
 & Conv&Conv&Conv&Conv&Conv&‌Full&‌Full&‌Full&&\\

CNN-M &$96\times 7\times 7$& $256\times 5\times 5$&$512\times 3\times 3$&$512\times 3\times 3$&$512\times 3\times 3$&4096&4096&1000&-&-\\
2014&  Stride 2& Stride 2&Stride 1&Stride 1&Stride 1&drop out&drop out &soft max&&\\
&LRN, x2 Pool&LRN, x2 Pool& -&-&x2 Pool&&&&&\\
\hline
 & Conv&Conv&Conv&Conv&Conv&‌Full&‌Full&‌Full&&\\

 CNN-S&$96\times 7\times 7$& $256\times 5\times 5$&$512\times 3\times 3$&$512\times 3\times 3$&$512\times 3\times 3$&4096&4096&1000&-&-\\
2014&  Stride 2& Stride 1&Stride 1&Stride 1&Stride 1&drop out&drop out &soft max&&\\
&LRN, x3 Pool& x2 Pool& -&-&x3 Pool&&&&&\\
\hline
\hline
& Layer 1&Layer 2&Layer 3&Layer 4&Layer 5&Layer 6&Layer 7&Layer 8&Layer 9&Layer 10\\
\cline{2-11}
 & Conv&Conv&Conv&Conv&Conv&Conv&Conv&Conv&Conv&Conv\\

 &$64\times 3\times 3$& $64\times 3\times 3$&$128\times 3\times 3$&$128\times 3\times 3$&$256\times 3\times 3$&$256\times 3\times 3$&$256\times 3\times 3$&$256\times 3\times 3$&$512\times 3\times 3$&$512\times 3\times 3$\\
&  Stride 1& Stride 1&Stride 1&Stride 1&Stride 1&Stride 1&Stride 1 &Stride 1&Stride 1&Stride 1\\
Very Deep& -& x2 Pool& -&x2 Pool&-&-&-&x2 Pool&-&-\\
\cline{2-11}
2014& Layer 11&Layer 12&Layer 13&Layer 14&Layer 15&Layer 16&Layer 17&Layer 18&Layer 19 &-\\
\cline{2-11}
& Conv&Conv&Conv&Conv&Conv&Conv&‌Full&‌Full&‌Full&\\

 &$512\times 3\times 3$&$512\times 3\times 3$&$512\times 3\times 3$&$512\times 3\times 3$&$512\times 3\times 3$&$512\times 3\times 3$&4096&4096&1000&-\\
&  Stride 1& Stride 1&Stride 1&Stride 1&Stride 1&Stride 1&drop out &drop out&soft max&\\
& -& x2 Pool& -&-&-&x2 Pool&&&&\\
\hline
\end{tabular}
}
\end{table*}

\subsection*{Shallow models}
\paragraph{HMAX model}
The HMAX model~\cite{serre2007feedforward} has a hierarchical architecture, largely inspired by the simple to complex cells hierarchy in the primary visual cortex proposed by Hubel and Wiesel~\cite{hubel1962receptive,hubel1968receptive}. The input image is first processed by the S1 layer (first layer) which extracts edges of different orientations and scales. Complex C1 units pool the outputs of S1 units in restricted neighborhoods and adjacent scales in order to increase position and scale invariance. Simple units of the next layers, including S2, S2b, and S3, integrate the activities of retinotopically organized afferent C1 units with different orientations. The complex units  C2, C2b, and C3 pool over the output of the corresponding simple layers, using a max operation, to achieve a global position and scale invariance. The employed HMAX model is implemented by Jim Mutch et. al. ~\cite{MUTCH10} and it is freely available at \url{http://cbcl.mit.edu/jmutch/cns/hmax/doc/}.
\paragraph{Pixel representation}
Pixel representation is simply constructed by vectorizing the gray values of all the pixels of an image. Then, these vectors are given to a linear SVM classifier to do the categorization. 

\subsection*{Image generation}
\begin{figure*}[!htb]
\centering
\includegraphics[scale=.7]{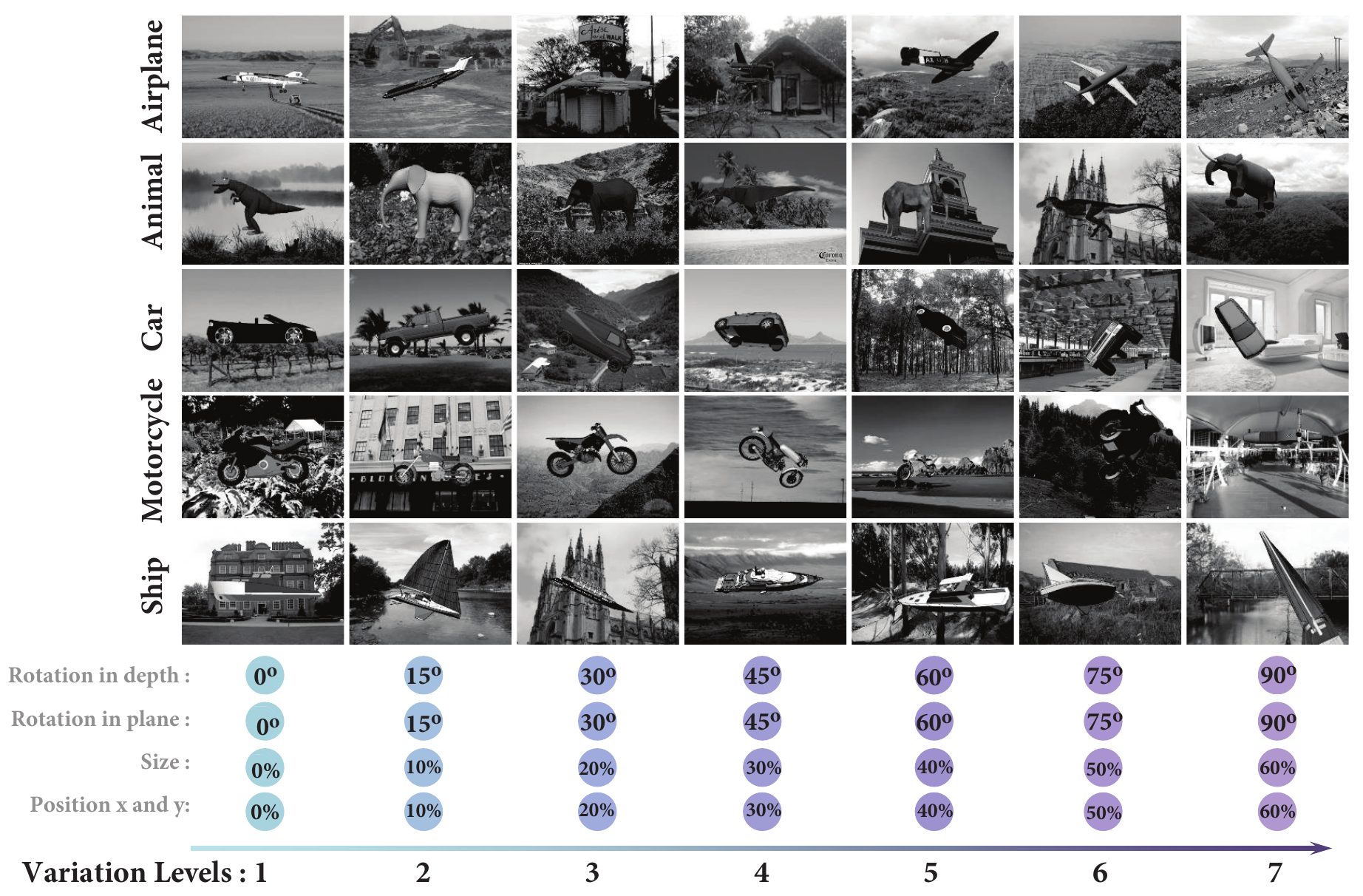}
\caption{{\bf Sample object images from the database superimposed on randomly selected natural backgrounds.} There are five object categories, each divided into seven levels of variations. Each 2-D image was rendered from a 3-D computer model. There were, on average, 16 various 3-D computer models for each object category. Objects vary in five dimensions: size, position (x, y), rotation in-depth,  rotation in plane, and background. To construct each 2-D image, we first randomly sampled from five different uniform distributions, each corresponding to one dimension. Then, these values were applied to the 3-D computer model, and a 2-D image was then generated. Variation levels start from no variations (Level 1, first column at left; note the values on horizontal axis) to high variation (Level 7, last column at right). For half of the experiments, objects were superimposed on  randomly selected natural images from a large pool of natural images (3,900 images), downloaded from the web. }
\label{fig1}
\end{figure*}

All  models were evaluated using an image database divided into five categories (airplane, animal, car, motorcycle, and ship) and seven levels of variations~\cite{ghodrati2014feedforward} (see Fig.~\ref{fig1}). The process of image generation is similar to Ghodrati et. al.~\cite{ghodrati2014feedforward}. Briefly, we built object images with different variation levels, where objects varied across five dimensions, namely: size, position (x and y), rotation in-depth, rotation in-plane, and background. To generate object images under different variations, we used 3-D computer models (3-D object images). Variations were divided into seven levels from no object variations (level 1) to mid- and high-level variations (level 7). In each level, random values were sampled from uniform distributions for every dimension. After sampling these random values, we applied them to the 3-D object model and generated a 2-D object image by snapshotting from the varied 3-D model. We performed the same procedure for all levels and objects. Note that the magnitude of variations in every dimension was randomly selected from uniform distributions that were restricted to predefined levels (i.e. from level 1 to 7). For example, in level three a random value between $0^{\circ}$ - $30^{\circ}$ was selected for in-depth rotation, a random value between $0^{\circ}$ - $30^{\circ}$ was selected for in-plane rotation, and so on (see Fig.~\ref{fig1}). The size of 2-D images were $300 \times 400$ pixels. As shown in Fig.~\ref{fig1}, for different dimensions, a higher variation level has broader variation intervals than the lower levels. There were on average 16 3-D image exemplars per category. All 2-D object images were then superimposed onto randomly selected natural images for experiment with natural backgrounds. There were over 3,900 natural images collected from the web, consisting of a variety of indoor and outdoor scenes.

\subsection*{Psychophysical experiments}
In total, 26 human subjects  participated in a  rapid invariant object categorization task (17 males and 9 females, age 21-32, mean age of 26 years). Each trial  started with a black fixation cross presented for 500 ms. Then an image was randomly selected from a pool of images and  was presented at the center of  screen for 25 ms (two frames, on a 80 Hz monitor). The image was  followed by a uniform blank screen presented for 25 ms, as an inter-stimulus interval (ISI). Immediately afterwards, a 1/f noise mask was presented for 100 ms to account for feed-forward processing and minimize the effects of back projections from higher visual areas. This type of masking is well established to be used in rapid object recognition tasks~\cite{cauchoix2016fast,ghodrati2014feedforward,crouzet2011visual,serre2007feedforward,lamme2002masking}. Finally, subjects had to select one out of five different categories using five keys, labeled on the keyboard. The next trial  started immediately after the key press. Stimuli were presented using MATLAB Psychophysics Toolbox~\cite{brainard1997psychophysics}  in a 21" CRT monitor with a resolution of $ 1024 \times 724 $ pixels,  a frame rate of 80 Hz, and viewing distance of 60 cm. Each stimulus covered $10^{\circ} \times 11^{\circ}$ of visual angle. Subjects were instructed to respond as fast and accurately as possible. All subjects voluntarily accepted to participate in the experiment and gave their written consent. The experimental procedure was approved by the local ethic committee.

According to the ``interruption theory"~\cite{breitmeyer2006visual,lamme2002masking,lamme2000distinct}, the visual system processes stimuli sequentially, so processing of a new stimulus (the noise mask) will interrupt the processing of the previous stimulus (the object image) before it can be modulated by the feedback signals from higher areas~\cite{lamme2002masking}. In our experiment, there is a  50 ms Stimulus Onset Asynchrony (SOA) between the object image and the noise mask (25 ms for image presentation and 25 ms for ISI). This SOA can disrupt IT-V4 ($\sim 40-60$ ms) and IT-V1 ($\sim 80-120$ ms) feedback signals, while it leaves the feed-forward information sweep intact~\cite{serre2007feedforward}. Using Transcranial
Magnetic Stimulation~\cite{lamme2000distinct}, it has been shown that applying magnetic pulses between 30 to 50 ms after stimulus onset will disturb the feed-forward visual information processing in the visual cortex. Thus, SOAs shorter than 50 ms would make the categorization task much harder by interrupting the feed-forward information flow.

Experiments were held in two sessions: in the first one, the objects were presented with a uniform gray background, and in the second one, a randomly selected natural background was used. Some subjects completed two sessions while others only participated in one session, so that each session was performed by 16 subjects. Each experimental session consisted of four blocks; each one containing 175 images (in total 700 images; 100 images per variation level, 20 images from each object category in each level). Subjects could rest between blocks for 5-10 minutes. Subjects performed a few training trials before starting the actual experiment (none of the images in these trials were presented in the main experiment). A feedback was shown to subjects during the training trials, indicating whether they responded correctly or not, but not during the main experiment.  
\subsection*{Model evaluation}
\textbf{Classification accuracy:} To evaluate the classification accuracy of the models, we first randomly selected 600 images from  each object category, variation level, and background condition (see Image generation section). Hence, we have 14 different datasets (7 variation levels $\times$ 2 background conditions), each of which consists of 3000 images (5 categories $\times$ 600 images). To compute the accuracy of each DCNN for a given variation level and background condition,  we randomly selected two subsets of 1500 training (300 images per category) and 750 testing images (150 images per category) from the corresponding image dataset. We then fed the pre-trained DCNN with the training and testing images and calculated the corresponding feature vectors for all layers. Afterwards, we used these feature vectors to train the classifier and compute the recognition accuracy of each layer. Here we used a linear SVM classifier (libSVM implementation~\cite{CC01a}, \url{www.csie.ntu.edu.tw/~cjlin/libsvm}) with  optimized regularization parameters. This procedure was repeated for 15 times (with different randomly selected training and testing sets) and the average and standard deviation of the accuracy were computed. This procedure was done for all models, levels, and layers.

For the HMAX and Pixel models, we first randomly selected 300 and 150 images (from each category and each variation level) as the training and testing sets, and then, computed their corresponding features. The visual prototypes of the S2, S2b and S3 layers of the HMAX model were randomly extracted from the training set, and the outputs of C2, C2b, and C3 layers were used to compute the performance of the HMAX model. Pixel representation for each image is simply a vector of pixels' gray values. Finally, the feature vectors were applied to a linear SVM classifier. The reported accuracies  are the average of 15 independent random runs.   

\noindent \textbf{Confusion matrix:} We also computed the confusion matrices for models and humans in all variation levels, both for objects on uniform and natural backgrounds. A confusion matrix allows us to determine which categories are more misclassified and how classification errors are distributed across different categories. For the models, confusion matrices were calculated from the labels assigned by the SVM. To obtain the human confusion matrix, we averaged the confusion matrices of all human subjects. 
\subsection*{Representational dissimilarity matrix (RDM)}
\textbf{Model RDM:} RDM provides a useful and illustrative tool to study the representational geometry of the response  to different images, and checking  whether images of the same category generate similar responses in the representational space. Each element in a RDM shows the pairwise dissimilarity between the response patterns elicited by two images. Here these dissimilarities are measured using Spearman's rank correlation distance (i.e., $1-$correlation). Moreover, RDMs is a useful tool to compare different representational spaces with each other. Here, we used RDMs to compare the internal representations of the models with human behavioral responses (see below). To calculate the RDMs, we used the RSA toolbox developed by Nili et.~al.~\cite{nili2014toolbox}.

\noindent \textbf{Human RDM:} Since we did not have access to the human internal object representations in our psychophysical experiment, we used the human behavioral scores to compute the RDMs (See~\cite{ghodrati2014feedforward} for more details). Actually, for each image, we computed the relative frequencies with which the image is assigned to different categories by all human subjects. Hence, we have a five-element vector for each image, which is used to construct the human RDM. Although, computing human RDMs based on behavioral responses is not a direct measurement of the representational content of the human visual system, it provides a way to compare internal representations of DCNN models to behavioral decisions of humans. 
 \begin{figure}[!htb]
\centering
\includegraphics[scale=.8]{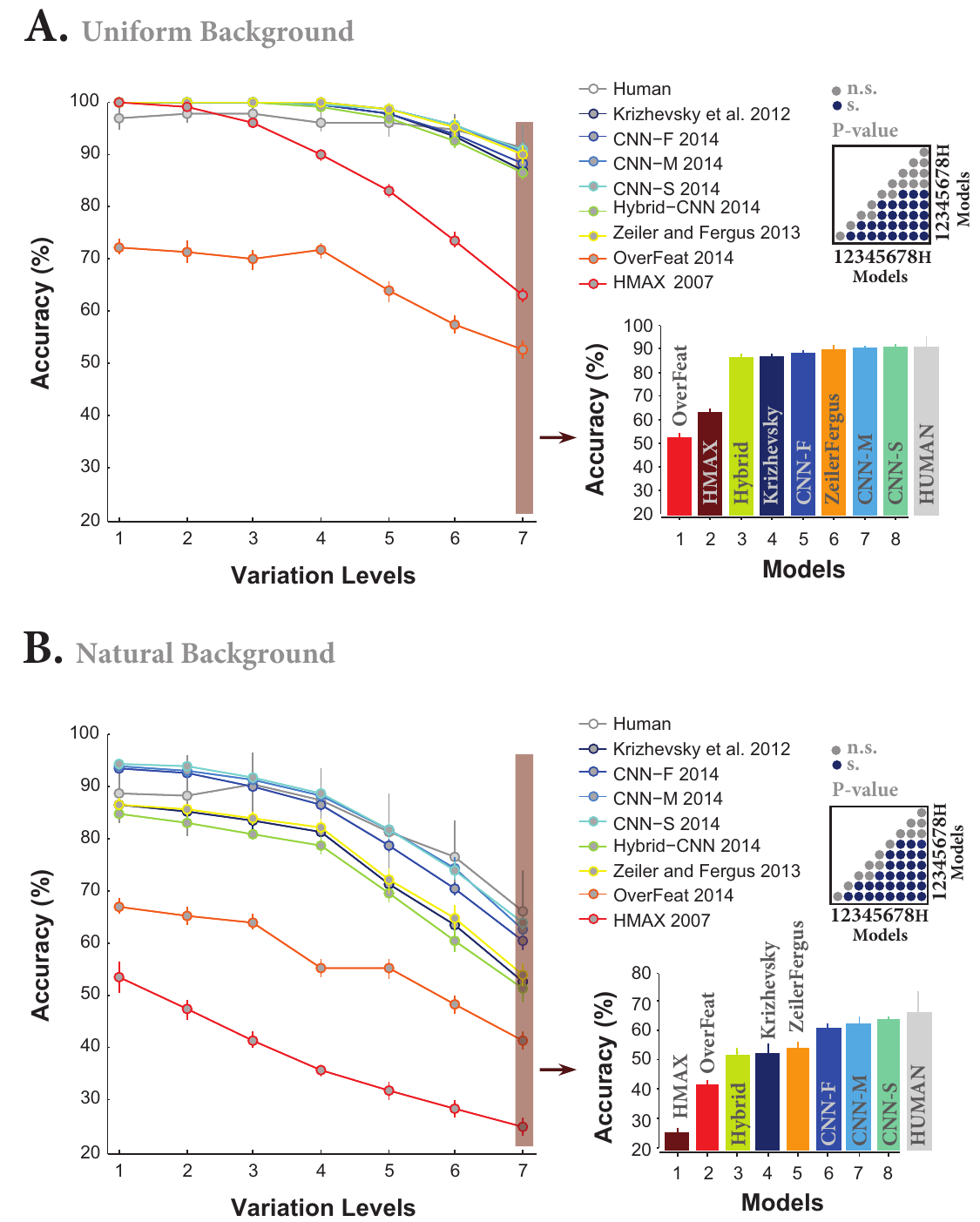}
\caption{{\bf Classification accuracy of models and humans in multiclass invariant object categorization task across seven levels of object variations.} A. Accuracies when objects were presented on uniform backgrounds. Each colored curve shows the accuracy of one model (specified in the legend). The gray curve indicates human categorization accuracy across seven levels. All models were well above chance level (20\%). The right panel shows the accuracies of both models and humans at the last level of variations (level seven; specified with pale, red rectangular), in ascending order. Level seven is considered  the most difficult level as the variations are high at this level, making the categorization difficult for models and human. The color-coded matrix, at the top-right of the bar plot, exhibits the p-values for all pairwise comparisons between human and models computed using the Wilcoxon rank sum tests. For example, the accuracy of the Hybrid-CNN was compared to the human and all other models and the pairwise comparison provides us with a p-value for each comparison. Blue points indicate that the accuracy difference is significant while gray points show insignificant differences. Numbers, written around the p-value matrix, correspond to models (H stands for human). Accuracies are reported as the average and standard deviation of 15 random, independent runs. B. Accuracies when objects were presented on randomly selected natural backgrounds.} 
\label{fig2}
\end{figure}
\section*{Results}
We tested the DCNNs in our invariant object categorization task including five object categories, seven variation levels, and two background conditions (see Materials and methods). The categorization accuracy of these models were compared with those of human subjects, performing rapid invariant object categorization tasks on the same images. For each model, variation level, and background condition, we randomly selected 300 training images and 150 testing ones per object category from the corresponding image dataset. The accuracy was then calculated over 15 random independent runs and the average and standard deviation were reported. We also analyzed the error distributions of all models and compared them to those of humans. Finally, we compared the representational geometry of models and humans, as a function of the variation levels.

\subsection*{DCNNs achieved human-level accuracy}
We compared the classification accuracy of the final layer of all models (DCNNs, and HMAX representation) with those of human subjects doing the invariant object categorization tasks in all variation levels and background conditions. Figure~\ref{fig2}A shows that almost all DCNNs achieved human-level accuracy across all levels when objects had a uniform gray background. The accuracies of DCNNs are even better than humans at low (levels 1 to 3) and intermediate (levels 4 and 5) variation levels. This might be due to inevitable motor errors that humans made during the psychophysical experiment, meaning that subjects might have perceived the image but pressed a wrong key. Also, it can be seen that the accuracies of humans and almost all DCNNs are virtually flat across all variation levels which means they are able to invariantly classify objects with uniform background.  Surprisingly, the accuracy of Overfeat is far  below the human-level accuracy, even worse than the HMAX model. This might be due to the structure and the number of features extracted by the Overfeat model which leads to a more complex feature space with high redundancy.

We compared the accuracy of humans and models at the most difficult level (7). There is no significant difference between the accuracies of CNN-S, CNN-M, Zeiler and Fergus, and human at this variation level (Fig.~\ref{fig2}A, bar plot; Also, see pairwise comparisons shown using a p-value matrix computed by the Wilcoxon rank sum test). CNN-S is the best model.

When we presented object images superimposed on natural backgrounds, the accuracies decreased for both humans and models. Figure~\ref{fig2}B illustrates that only three DCNNs (CNN-F, CNN-M, CNN-S) performed close to human. The accuracy of the HMAX model dropped down just above chance level (i.e., 20\%) at the seventh variation level. Interestingly, the accuracy of Overfeat remained almost constant either in objects on uniform or natural backgrounds, suggesting that this model is more suitable for tasks with unsegmented images. Similarly, we compared the accuracies at the most difficult level (level 7) when objects had natural backgrounds. Again, there is no significant difference between the accuracies of CNN-S, CNN-M, and humans (see the p-value matrix computed using  the Wilcoxon rank sum test for all possible pairwise comparisons). However, the accuracy of human subjects is significantly above the HMAX model and other DCNNs (i.e., CNN-F, Zeiler and Fergus, Krizhevsky, Hybrid-CNN, and Overfeat).

\subsection*{How accuracy evolves across layers in DCNNs}
DCNNs have a hierarchical structure of different processing stages in which each layer extracts a large pool of features (e.g., $>4000$ features at top layers). Therefore, the computational load of such models is very high. This raises important questions: what is the contribution of each layer to the final accuracy? and how does the accuracy evolve across the layers?

\begin{figure*}[!htb]
\centering
\includegraphics[scale=0.8]{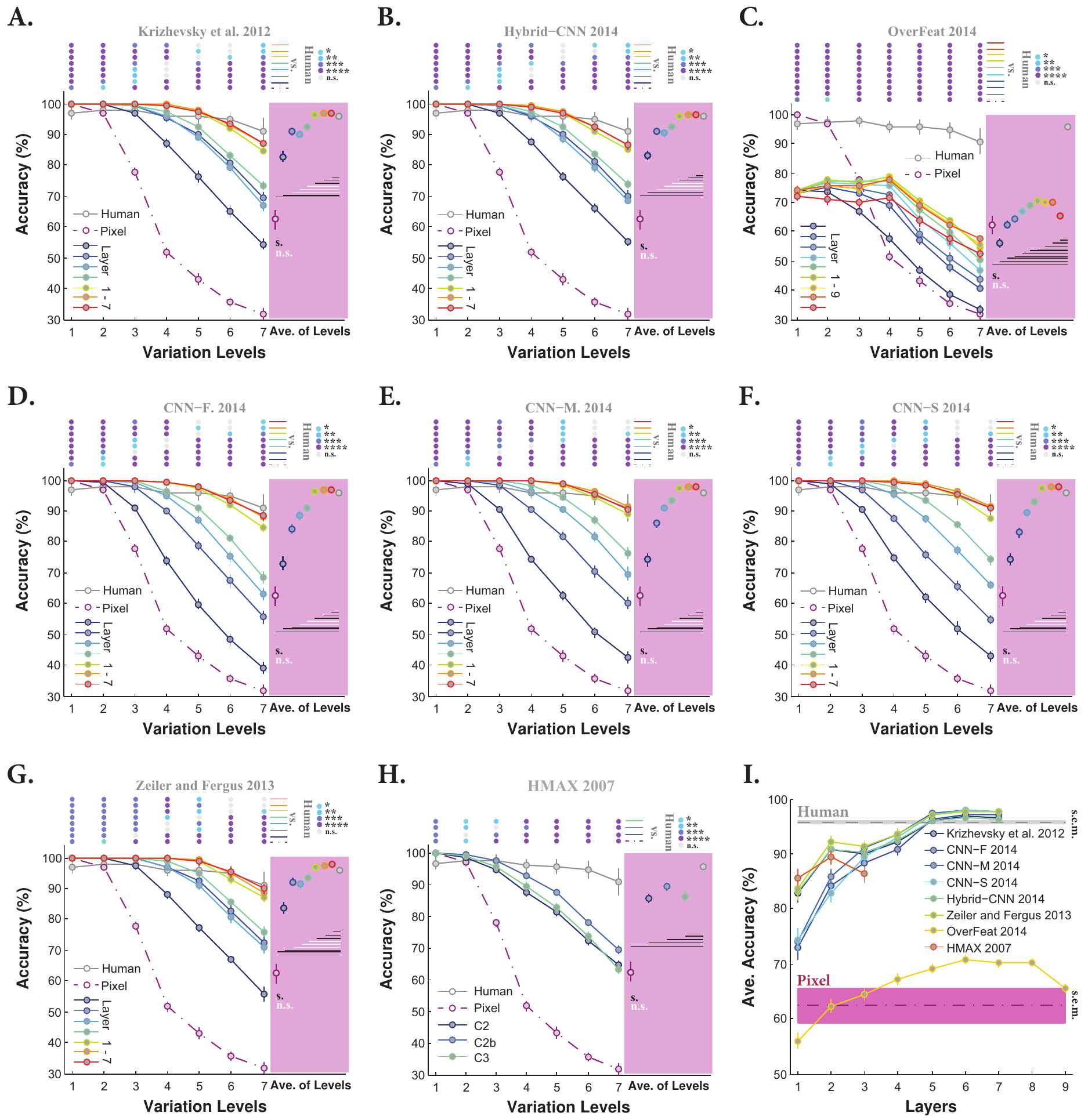}
\caption{{\bf Classification accuracy of models (for all layers separately) and humans in multiclass invariant object categorization task across seven levels of object variations, when objects had uniform backgrounds.} A. Accuracy of Krizhevsky et. al. 2012 across all layers and levels. Mean accuracies and s.e.m. are reported  using 15 random, independent runs. Each colored curve shows the accuracy of one layer of the model (specified on the bottom-left legend). The accuracy of Pixel representation is depicted using a dashed, dark purple curve. The gray curve indicates human categorization accuracy across seven levels. The chance level is 20\%; no layer hit the chance level for this task (note that the accuracy of Pixel representation dropped down to 10\% above chance at level seven). The color-coded points at the top of the plot indicate whether there is a significant difference between the accuracy of humans and model layers (computed using the Wilcoxon rank sum test). Each color refers to a p-value, specified on the top-right ($*$: $p<0.05$, $**$: $p<0.01$, $***$: $p<0.001$, $****$: $p<0.0001$). Colored circles on the pink area, show the average accuracy of each layer, across all variation levels (one value for each layer and all levels), with the same color code as curves. The horizontal lines, depicted underneath the circles, indicate whether the difference between human accuracy (gray circle) and layers of the model is significant (computed using the Wilcoxon rank sum test; black line: significant, white line: insignificant). B-H. Accuracies of Hybrid-CNN, Overfeat, CNN-F, CNN-M, CNN-S, Zeiler and Fergus, and HMAX model, respectively. I. The average accuracy across all levels for each layer of each model (again error bars are s.e.m.). Each curve corresponds to a model. This simply summarizes the accuracies, depicted in the pink areas. The shaded area shows the average baseline accuracy (pale-purple, Pixel representation) and human accuracy (gray) across all levels.}
\label{fig3}
\end{figure*}

We addressed these questions by calculating the accuracy of each layer of the models across all variation levels. This provides us with the contribution of each layer to the final accuracy. Figure~\ref{fig3}A-H shows the accuracies of all layers and models when objects had uniform gray background. The accuracies of the Pixel representation (dashed, dark purple curve) and human (gray curve) are also shown on each plot. 

\begin{figure*}[!htb]
\centering
\includegraphics[scale=0.8]{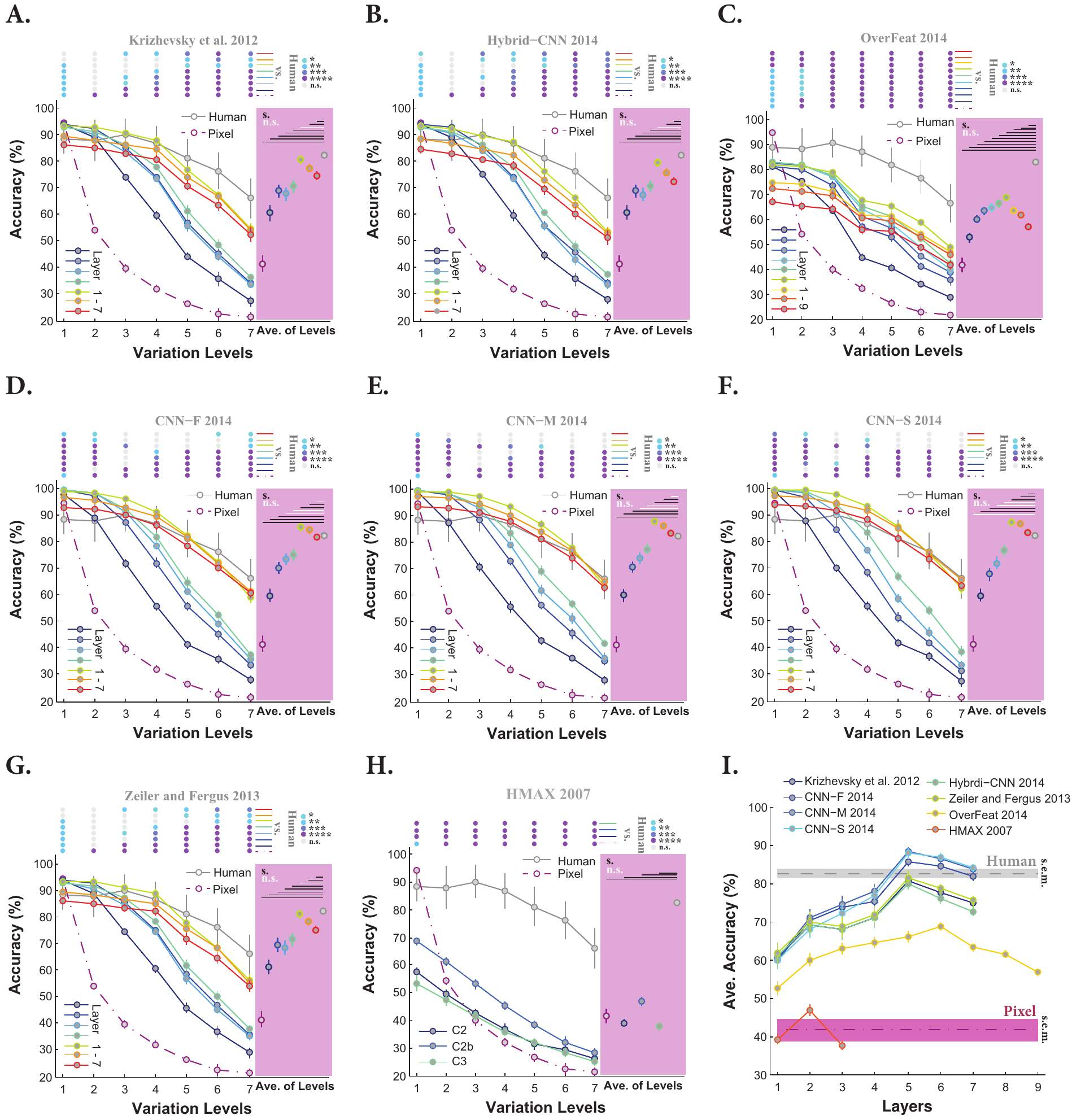}
\caption{{\bf Classification accuracy of models (for all layers separately) and human in multiclass invariant object categorization task across seven levels of object variations, when objects had natural backgrounds.} A-H. Accuracies of Krizhevsky et. al., Hybrid-CNN, Overfeat, CNN-F, CNN-M, CNN-S, Zeiler and Fergus, and HMAX model across all layers and variation levels, respectively. I. The average accuracy across all levels for each layer of each model (again error bars are s.e.m.). Details of diagrams are explained in the caption of Fig.~\ref{fig3}}
\label{fig4}
\end{figure*}

Overall, the accuracies  significantly evolved across layers of DCNNs. Moreover, almost all layers of the models (except Overfeat), even Pixel representation, achieved perfect accuracies at low variation levels (i.e., levels 1 and 2), suggesting that this task is very simple when objects had small variations and uniform gray background. Looking at the intermediate and difficult variation levels shows that the accuracies tend to increase as we go up across the layers. However, the trend is different between layers and models. For example, layers 2, 3, and 4 in three DCNNs (Krizhevsky, Hybrid-CNN, Zeiler and Fergus) have very similar accuracies across the variation levels (Fig.~\ref{fig3}A, B, and G). Similar results can be seen for these models in layers 5, 6, and 7 (Fig.~\ref{fig3}A, B, and G). In contrast, there is a high increase in accuracies from layer 1 to 4 for CNN-F, CNN-M, and CNN-S, while the three last layers have similar accuracies. There is also a gradual increase in the accuracy of Overfeat from layer 2 to 5 (with the similar accuracy for layers 6, 7, and 8); however, there is a considerable decrease at the output layer (Fig.~\ref{fig3}C).  Moreover, the overall accuracy of Overfeat is low compared to humans and other models as previously seen in Fig.~\ref{fig2}.

Interestingly, the accuracy of  HMAX, as a shallow model, is far below the accuracies of DCNNs  (C2b is the best performing layer). This shows the important role of supervised deep learning in achieving high classification accuracy. As expected, the accuracy of Pixel representation exponentially decreased down to 30\% at level seven, confirming the fact that invariant object recognition requires multi-layered architectures (note that the chance level accuracy is 20\%). We note, however, that Pixel performs very well with no viewpoint variations (level 1).

We also compared the accuracies of all layers of the models with those of humans. Color-coded points at the top of each plot in Fig.~\ref{fig3} indicate the p-values of the Wilcoxon rank sum test.  The average accuracy of each layer across all variation levels is shown on the pink area at the right side of each plot, summarizing the contribution of each layer to final accuracy independently of variation levels. Horizontal lines on the pink area show whether the average accuracy of each layer is significantly different from those of humans (black: significant; white: insignificant). Furthermore, Fig.~\ref{fig3}I summarizes the results depicted on the pink areas, confirming that the last three layers in DCNNs (except Overfeat) have similar accuracies. 

\begin{figure*}[!htb]
\centering
\includegraphics[scale=0.7]{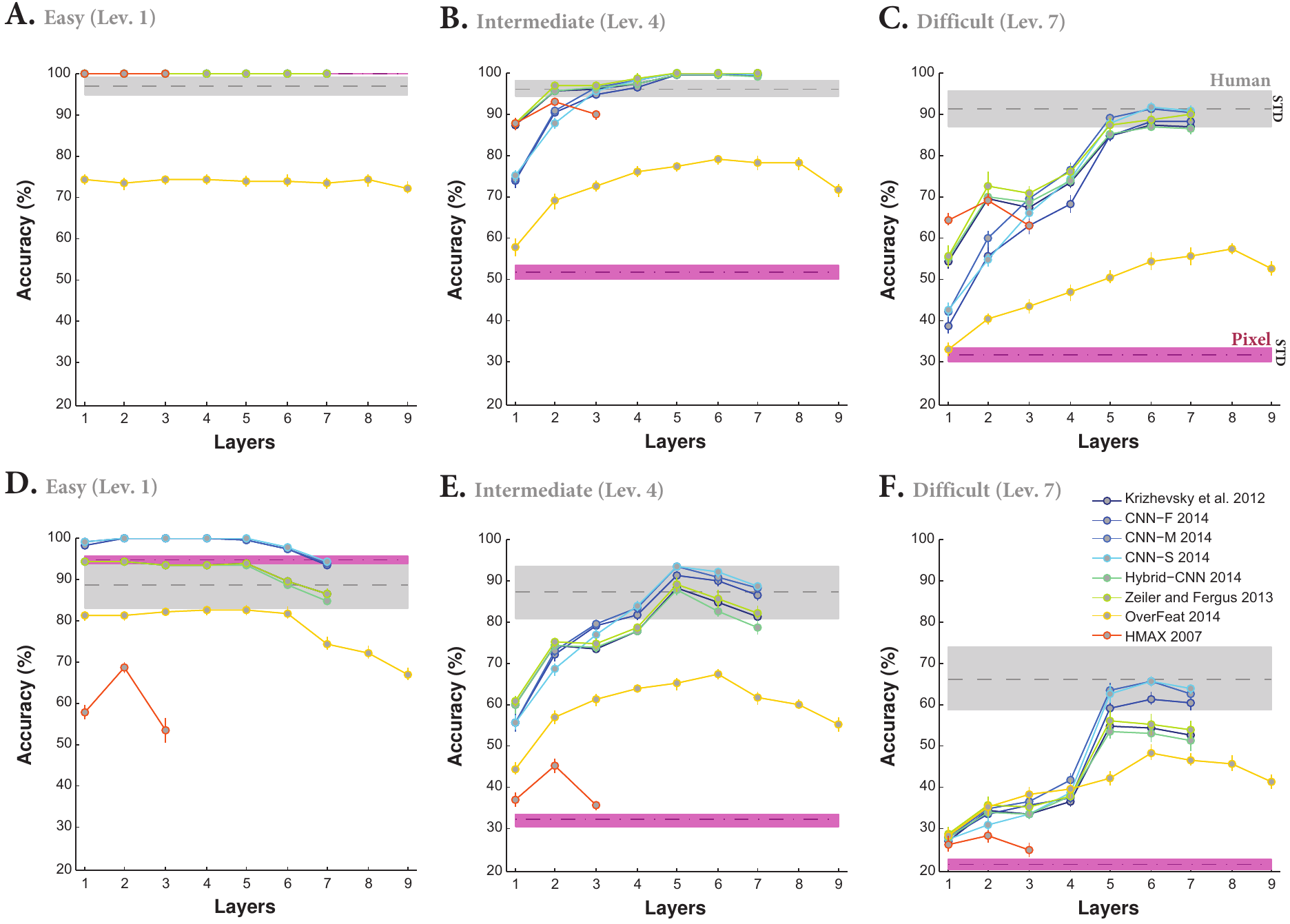}
\caption{{\bf Classification accuracy at easy (level 1), intermediate (level 4) and difficult (level 7) levels for different layers of the models.} A-C. Accuracy for different layers at easy (A), intermediate (B) and difficult (C) levels when objects had uniform backgrounds. Each curve represents the accuracy of a model. The shaded areas show the accuracy of the Pixel representation (pale purple) and human (gray). Error bars are standard deviation. D-F. Idem when objects had natural backgrounds.}
\label{fig5}
\end{figure*}

We also  tested the models on objects with natural backgrounds to see whether the contributions of similarly performing layers change in more challenging tasks. Not surprisingly, the accuracy of human subjects dropped by ~10\% at low variation level (level 1), and down to ~25\% at high variation level (level 7) with respect to the uniform background case (Fig.~\ref{fig4}, gray curve). Not surprisingly, the Pixel representation shows an exponential decline in the accuracy across the levels, with the chance accuracy at level seven (Fig.~\ref{fig4}, dashed dark purple curve). Similar to Fig.~\ref{fig3}, all DCNNs, excluding Overfeat, achieved close to human-level accuracy at low variation levels (levels 1, 2, and 3). Interestingly, the Pixel representation performed better than most models at level one, suggesting that object categorization at low variation level can be done without elaborate feature extraction methods (note that we had only five object categories, therefore, this can be different with more categories).

The severe drop in the accuracy of the HMAX model with respect to the uniform background experiment reflects the difficulty of this model to cope with distractors in natural backgrounds. For both background conditions, the C2b layer has higher accuracy than C3 layer and can better tolerate object variations. The main reason why HMAX is not performing as well as DCNNs is probably the lack of a purposive learning rule~\cite{kheradpisheh2015bio,pinto2011comparing}. HMAX randomly extracts a large number of visual features (image crops) which could be highly redundant, uninformative, and even misleading~\cite{ghodrati2012can}. The issue of inappropriate features becomes more evident when the background is clutter. 

Another noticeable fact about DCNNs in the natural background experiment is the superiority of the last convolutional layers with respect to the fully connected layers; for example, the accuracy of the fifth layer in the Krizhevsky model is higher than the seventh layer's. One possible reason for the low accuracies in the final layers of DCNNs is that the fully connected layers are designed to perform classification themselves, and not to provide input for a SVM classifier. Besides, the fully connected layers were  optimized for Imagenet classification, but not for our dataset. A last reason could be that the convolutional layers have more features than the fully connected layers.

\begin{figure*}[!htb]
\centering
\includegraphics[scale=.8]{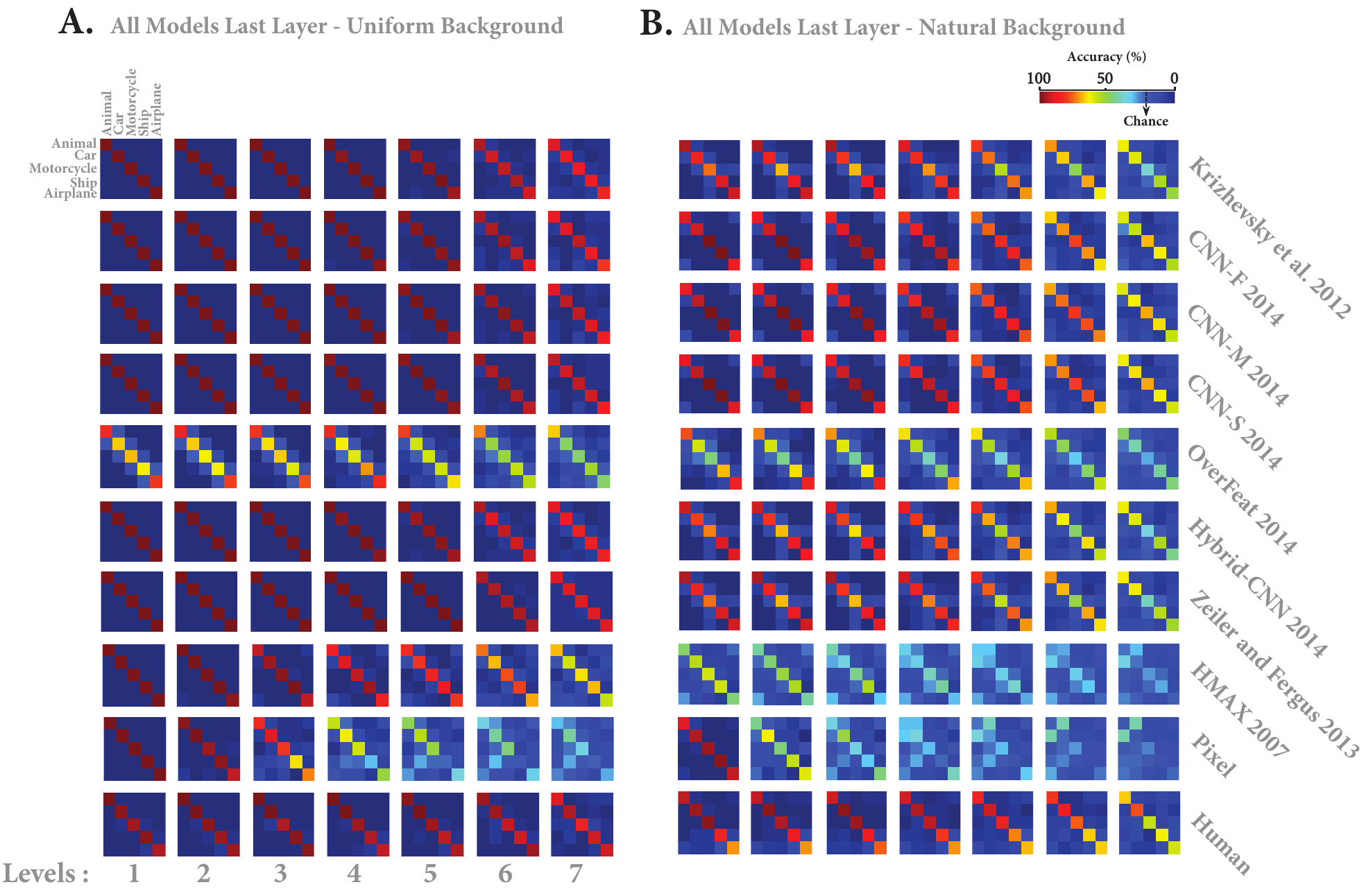}
\caption{{\bf Confusion matrices for multiclass invariant object categorization task.} A. Each color-coded matrix shows the confusion matrix of a model when categorizing different object categories (specified in the first matrix at the top-left corner), when images had uniform backgrounds. Each row corresponds to a model. Last row shows human confusion matrix. Each column indicates a particular level of variation (levels 1 to 7). Models' name is depicted at the right end. B. Idem with natural backgrounds. The color bar at the top-right shows the percentage of the labels assigned to each category, The chance level indicated with an arrow. Confusion matrices were calculated only for the last layers of the models.}
\label{fig6}
\end{figure*}

Given the accuracies of all layers, it can be seen that the accuracies  evolved across the layers. However, similar to Fig.~\ref{fig3}, layers 2, 3, and 4 of Krizhesvky, Zeiler and Fergus, and Hybrid-CNN contribute almost equally to the final accuracy. 
Again, CNN-F, CNN-M, and CNN-S showed a different trend in terms of the contribution of each layer to the final accuracy. Moreover, as shown in Fig.~\ref{fig4}D-F, only these three models achieved human-level accuracy at difficult levels (levels 6 and 7). The accuracies of other DCNNs, however, are significantly lower than humans at these levels (see the color-coded points in Fig.~\ref{fig4}A-C, G which indicate the p-values computed by the Wilcoxon rank sum tests). We summarized the average accuracies across all levels for each layer of the models, shown as color-coded circles with error bars on the pink areas next to each plot. In most cases, layer 5 (the last convolutional layer - layer 6 in Overfeat) has the highest accuracy among layers. This is summarized in Fig.~\ref{fig4}I, which is actually the summary of results shown on pink areas. Figure~\ref{fig4}I also confirms that only CNN-F, CNN-M, and CNN-S achieve human-level accuracy.

We further compared the accuracies of all layers of the models with humans at the easy (level 1), intermediate (level 4) and difficult (level 7) variation levels to see how each layer performs the task as the level of variations increases. Figure~\ref{fig5}A-C show the accuracies for the uniform background condition. The easy level is not very informative because of a ceiling effect: all models (but Overfeat) reach 100\% accuracy. At the intermediate level, all DCNNs (except Overfeat) reached  the human-level accuracy from layer 4 upwards (Fig.~\ref{fig5}A), suggesting that even with intermediate level of variation, DCNNs have remarkable accuracies (note that objects had uniform background). This is clearly not true for the HMAX and Overfeat networks. However, when models were fed with images from the most difficult level, only the last layers (layers 5, 6, and 7) achieved human-level accuracy (see Fig.~\ref{fig5}B). Notably the last three layers have almost similar accuracies.

When objects had natural backgrounds, somewhat surprisingly the accuracies of all DCNNs	 (but Overfeat) is maximal with layer 2, and drops for subsequent layers. This shows that deeper is not always better. The fact that the Pixel representation performs well at this level confirms this finding. At the intermediate level, the picture is different: only the last three layers of DCNNs, excluding Overfeat, reach human-level accuracy (see Fig.~\ref{fig5}E). Finally, at the seventh variation level, Figure~\ref{fig5}F shows that only three DCNNs reach human performance: CNN-F, CNN-M, and CNN-S.

In summary, the above results, taken together, illustrate that some DCNNs are as accurate as humans, even at the highest variation levels.

\subsection*{Do DCNNs and humans make similar errors?} 
The accuracies reported in the previous section only represent the ratio of correct responses. Indeed, they did not reflect whether  models and humans made similar misclassifications. To do a more precise and category-based comparison between the recognition accuracies of humans and models, we computed the confusion matrices for each variation level. Figure~\ref{fig6} provides the confusion matrices for humans and the last layers of all models for both uniform (see Fig.~\ref{fig6}A) and natural (see Fig.~\ref{fig6}B) backgrounds, and for each variation level.

\begin{figure*}[!htb]
\centering
\includegraphics[scale=0.8]{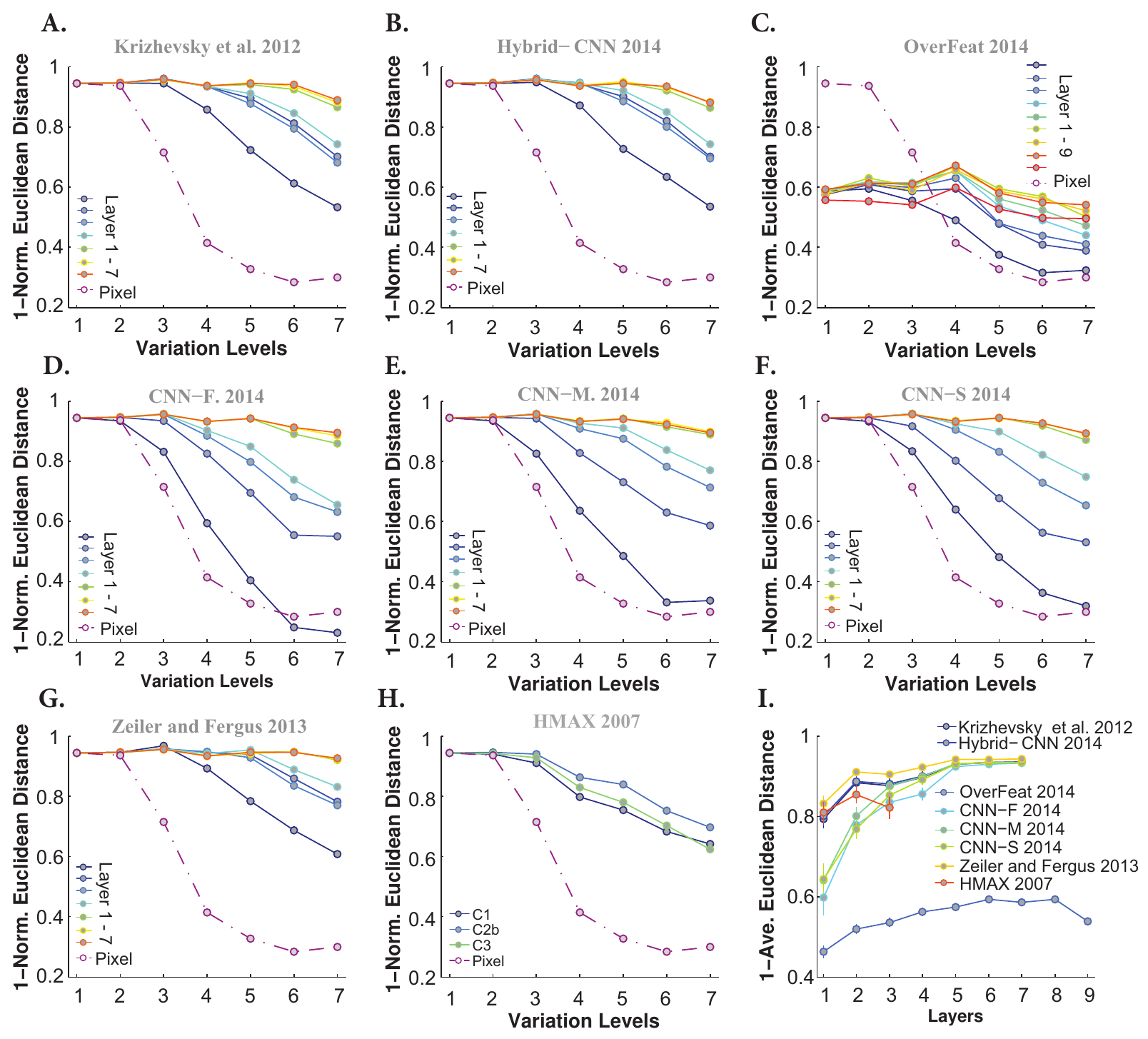}
\caption{{\bf Similarity between models' and humans' confusion matrices when images had uniform backgrounds.} A. Similarity between Krizhevsky et al. 2012 confusion matrices and that of humans (measured as 1-normalized Euclidean distance). Each curve shows the similarity between human confusion matrix and one layer of Krizhevsky et al. 2012 (specified on the right legend), across different levels of variations. The similarity between the confusion matrix of the Pixel representation and humans is shown using a dark purple, dashed line. B-H. Idem for the Hybrid-CNN, Overfeat, CNN-F, CNN-M, CNN-S, Zeiler and Fergus, and HMAX models, respectively. I. The average similarity across all levels for each layer of each model (error bars are s.e.m.). Each curve corresponds to one model.}
\label{fig7}
\end{figure*}

Despite a very short presentation time in the behavioral experiment, humans performed remarkably well at categorizing five object classes, either when object had uniform (Fig.~\ref{fig6}A, last row) or natural (Fig.~\ref{fig6}B, last row) backgrounds, with minimum misclassifications across different categories and levels. It is, however, important to point out that the majority of human  errors corresponded to ship - airplane confusions. This was probably  due to the shape similarity among these objects (e.g., both categories usually have bodies, sails, wings, etc.).
     
Figure~\ref{fig6} demonstrates that the HMAX model and Pixel representation misclassified almost all categories at high variation levels. With natural backgrounds, they uniformly assigned input images into different classes. Conversely, DCNNs show few classification errors across different categories and levels, though the distribution of errors is different from one model to another. For example, the majority of recognition errors made by Krizehvsy, Zeiler and Fergus, and Hybrid-CNN belonged to car and motorcycle classes, while animal and airplane classes were mostly misclassified by CNN-F, CNN-M, and CNN-S. Finally, Overfeat shows evenly-distributed errors across categories, confirming its low accuracy.  

\begin{figure*}[!htb]
\centering
\includegraphics[scale=0.8]{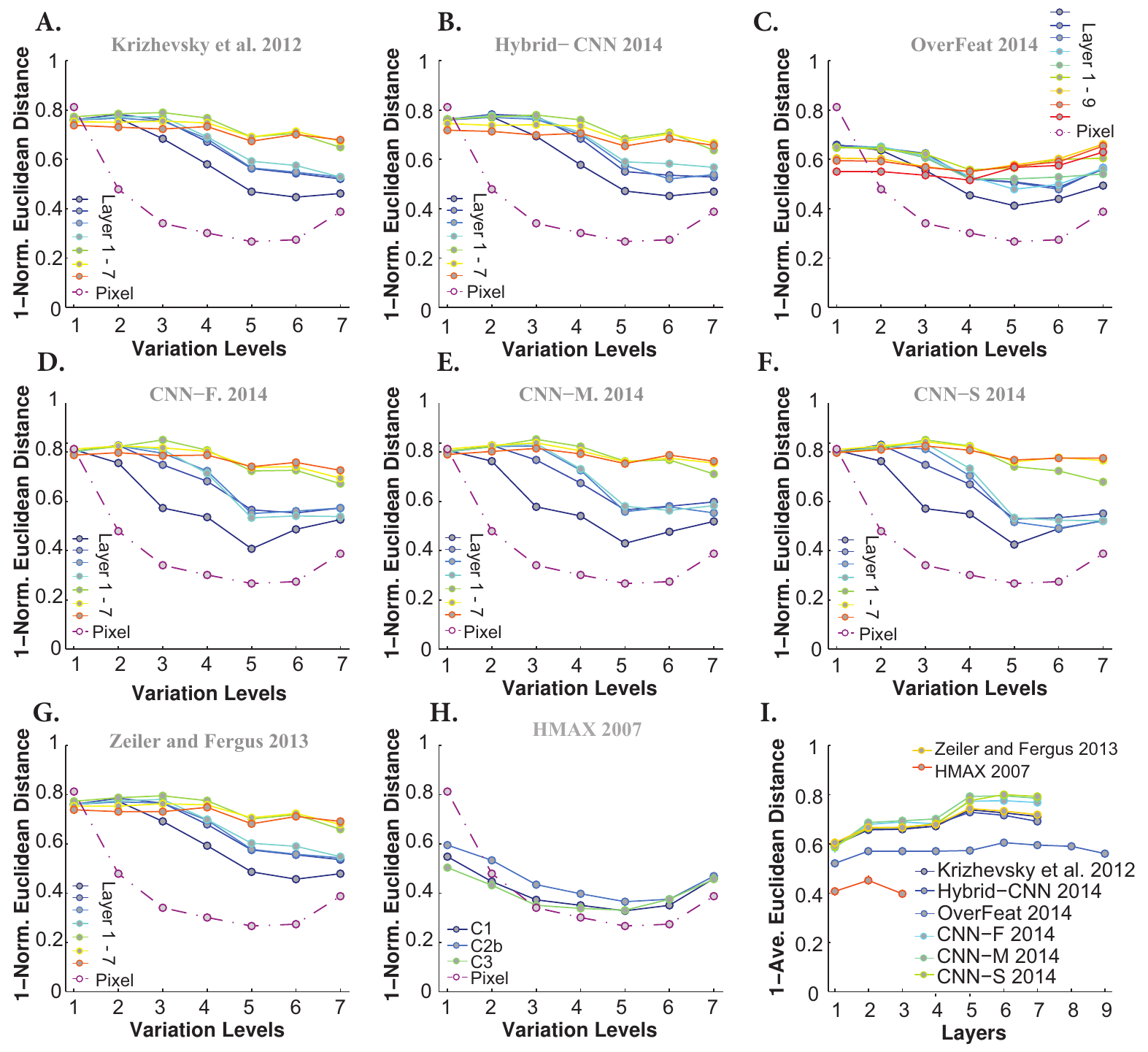}
\caption{{\bf Similarity between models' and humans' confusion matrices, when object images had natural backgrounds.} A-H. Similarities between the confusion matrices of Krizhevsky, Hybrid-CNN, Overfeat, CNN-F, CNN-M, CNN-S, Zeiler and Fergus, HMAX model and that of humans. Figure conventions are identical to Fig.~\ref{fig7}. I. The average similarity across all levels for each layer of each model (error bars are s.e.m.). Each curve corresponds to a model.}
\label{fig8}
\end{figure*} 

We also examined whether models' decisions are similar to those of humans. To this end, we computed the similarity between the humans' confusion matrices and those of the models. An important point is to factor out the impacts of the mean accuracies (of humans and models) on the similarity measure, to only take the error distributions into account. Therefore, for each confusion matrix, we first excluded the diagonal terms and arranged the remaining elements in a vector and normalized it by its \textit{L2} norm. Then, the similarity between two confusion matrices is computed using the Euclidean distance between their corresponding vectors subtracted from one (here we call it as \textit{1 - Norm. Euclidean distance}). In this way, we are just comparing the error distributions of humans and models independent of their accuracies. Figure~\ref{fig7} provides the similarities between models and humans across all layers and levels when objects had uniform background. Almost all models, including the Pixel representation, show the maximum possible similarity at low variation levels (levels 1 and 2). However, the similarity of Pixel representation exponentially decreases from level 2 upwards. Overall, the highest layers of DCNNs (except Overfeat) are more similar to humans' decisions. This point is also shown in Figure~\ref{fig7}I, which represents the average similarities across all variation levels (each curve corresponds to one model). Note that due to the high recognition accuracies in uniform background condition, this level of similarity was predictable. 

The similarity between models' and humans' errors, however, decreases in the case of images with natural backgrounds. The  HMAX model had the lowest similiarity with human (see Fig.~\ref{fig8}). Although DCNNs have reached human-level accuracy, their decisions and distribution of errors are different from human's. Interestingly, the Overfeat has almost a constant similarity across layers and levels. Comparing the similarities across DCNNs shows that CNN-F, CNN-M, and CNN-S have the highest similarities to humans, which is also reflected in Fig.~\ref{fig8}I.

\begin{figure*}[!htb]
\centering
\includegraphics[scale=0.8]{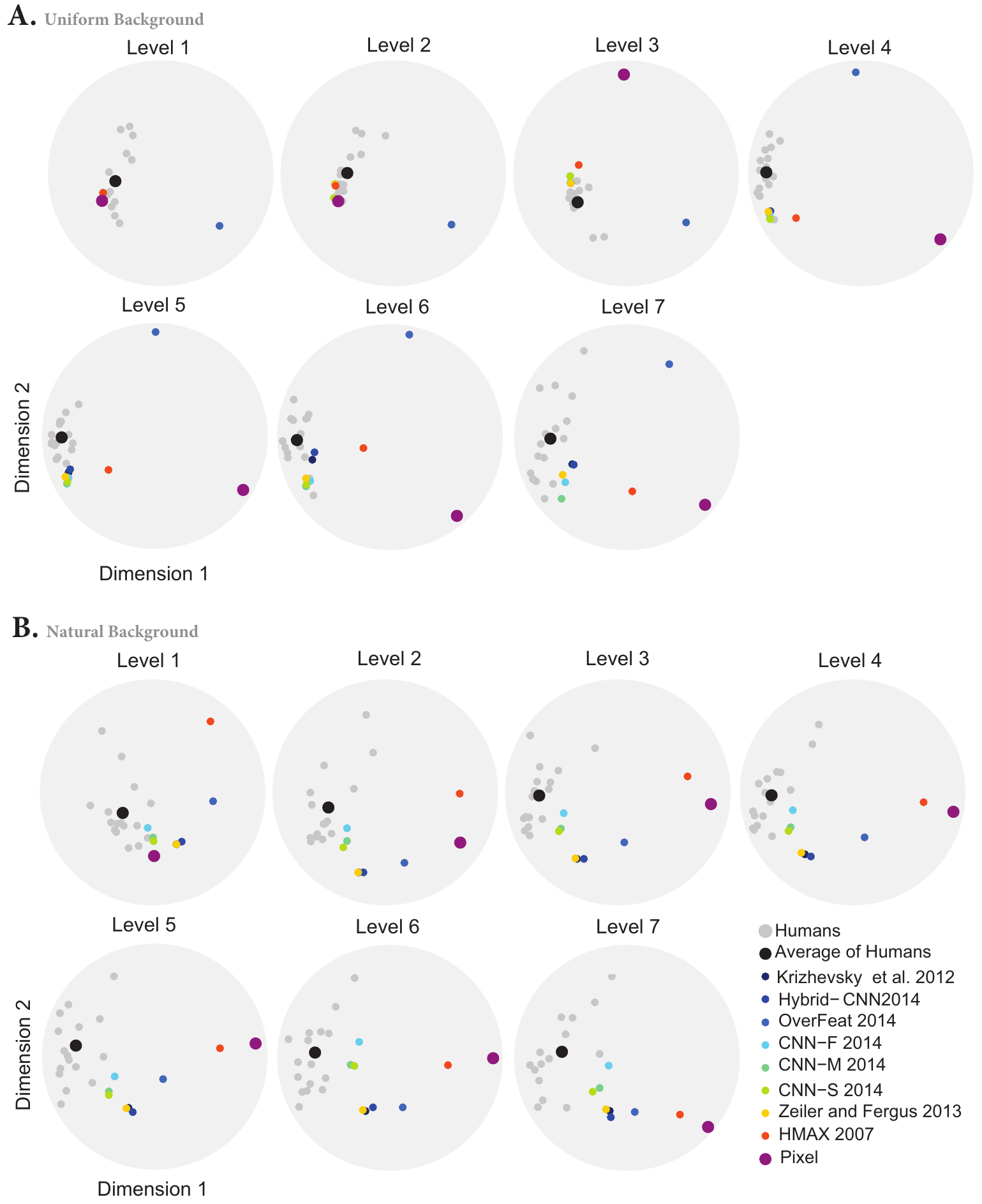}
\caption{\textbf{The distances between models and humans visualized using the multidimensional scaling (MDS) method.} distances between models and humans when images had uniform (A) and natural backgrounds (B). Light gray circles show the position of each human subjects and larger black circle shows the average of all subjects. Color circles represent models.}
\label{fig9}
\end{figure*}

To summarize our results so far: the best DCNNs can reach human performance even at the highest variation level, but their error distributions are different to the average human one (similarity $<$ 1 on Fig.~\ref{fig8}). However, one needs a reference here, because humans also differ between each other. Are these difference between humans smaller than differences between humans and DCNNs? To investigate this issue, we used the multidimensional scaling (MDS) method to visualize the distances (i.e., similarities) between the confusion matrices of humans and models (last layer) in 2-D maps (see Figure~\ref{fig9}). Each map corresponds to a certain variation level and background condition.

In the uniform background condition, humans have small inter-subject distances. As we move from low to high variations, the distance between DCNNs and humans becomes greater. In high variation levels, the Overfeat, HMAX, and Pixel models are very far from the human subjects as well as from the other DCNNs. The other models remain indiscernible from humans.

In the natural background condition, the human between-subject distances are relatively higher than in the uniform condition. As the level of variations increases, the models tend to get further away from the human subjects. But the CNN-F, CNN-M, and CNN-S are difficult, if not impossible, to discern from humans.

So far, we have analyzed the accuracies and error distributions of models and humans, when features were used by a SVM classifier. However, such analyses do not inform us about the internal representational geometry of models and their similarities to those of humans. It is very important to investigate how different categories are represented in the feature space. 

\subsection*{Representational geometry of models and human}

Representational similarity analysis has become a popular tool to study the internal representation of  models~\cite{khaligh2014deep,yamins2014performance,cadieu2014deep,kriegeskorte2008matching}  in response to different object categories. The representational geometries of models can then be compared with neural responses independently of the recording modality  (e.g. fMRI~\cite{kriegeskorte2008matching,khaligh2014deep}, cell recording~\cite{kiani2007object,yamins2014performance,cadieu2014deep}, behavior~\cite{carlson2014reaction,kriegeskorte2013representational,mur2013human,ghodrati2014feedforward}, and MEG~\cite{carlson2013representational}), showing to what degree each model resembles the brain representations. Here, we calculated representational dissimilarity matrices (RDM) for models and humans~\cite{nili2014toolbox}. We then compared the RDMs of humans and each model and quantified the similarity between these two. Model RDMs were calculated based on pairwise correlation between the feature vectors of two images (see Materials and methods). To calculate the human RDM, we used their behavioral scores recorded in the psychophysical experiment (see Materials and methods as well as~\cite{ghodrati2014feedforward}).

\begin{figure*}[!htb]
\centering
\includegraphics[scale=.8]{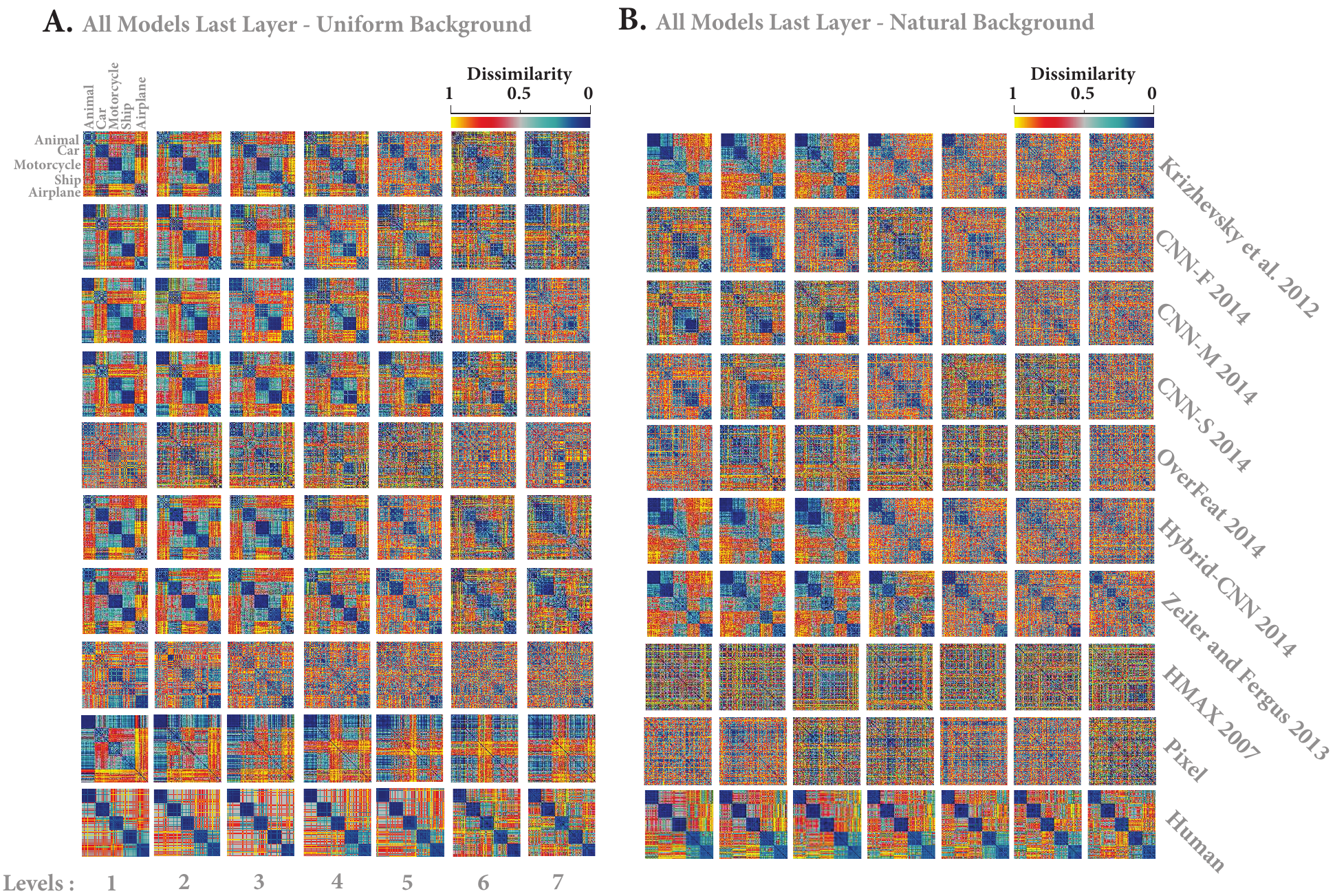}
\caption{{\bf Representational Dissimilarity Matrices (RDM) for models and humans.}  RDMs for humans and models when images had uniform (A) and natural (B) backgrounds. Each element in a matrix shows the pairwise dissimilarities between the internal representations of the two images (measured as $1-$ Spearman's rank correlation). Each row of RDMs corresponds to a model (specified on the right) and each column corresponds to a particular level of variation (from level 1 to 7). Last row illustrates the human RDMs, calculated from the behavioral responses. The color bar on the top-right corner shows the degree of dissimilarity. For the sake of visualization, we only included  20 images from each category, leading to $100\times 100$ matrices. Model RDMs were calculated for the last layer of each model.}
\label{fig10}
\end{figure*}

Figure~\ref{fig10} represents the RDMs for models and human across different levels of variation both for objects on uniform (Fig.~\ref{fig10}A) and natural (Fig.~\ref{fig10}B) backgrounds. Note that these RDMs are calculated from the object representations in the last layers of the models. For better visualization, we show only 20 images from each category; therefore, the size of RDMs is $100\times 100$ (reported RDMs were averaged over six random runs).

 As expected, human RDM clearly represents each object category, with minimum intra-class dissimilarity and maximum inter-class dissimilarity, across all variation levels (last row in Fig.~\ref{fig10}A  and Fig.~\ref{fig10}B for uniform and natural backgrounds, respectively). However, both HMAX  and Pixel representation show a random pattern in their RDMs when objects had natural backgrounds (Fig.~\ref{fig10}B, rows 8 and 9), suggesting that such low and intermediate visual features are unable to invariantly represent different object categories. The situation is slightly better when object had uniform background (Fig.~\ref{fig10}A, rows 8 and 9). In this case, there is some categorical information, mostly across low variation levels (levels 1 to 3, and 4 to some extent), for animal, motorcycle, and airplane images. Such information is attenuated at intermediate and high variation levels.
 \begin{figure*}[!htb]
\centering
\includegraphics[scale=0.8]{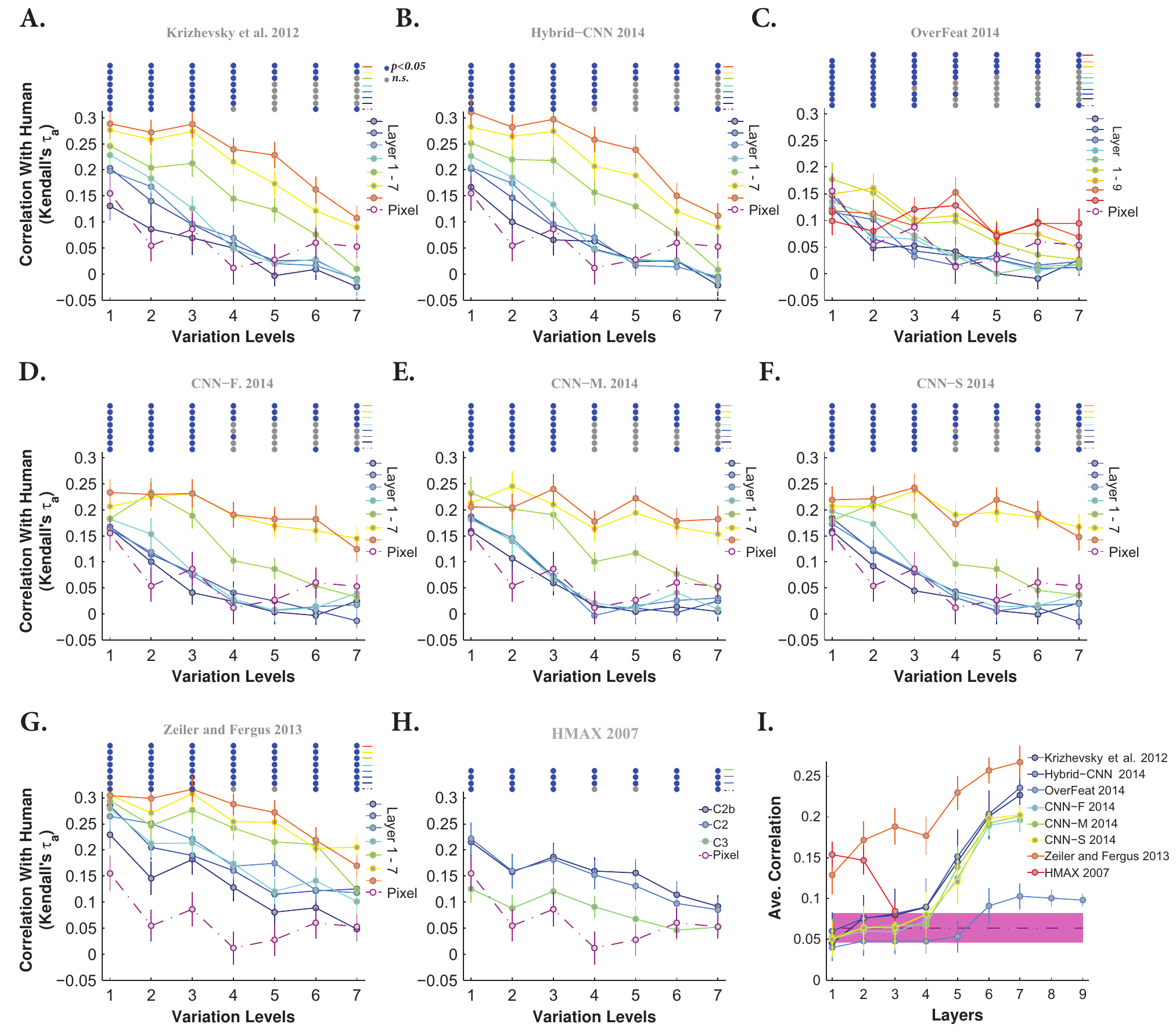}
\caption{{\bf Correlation between humans' and models' RDMs, across different layers and levels, when objects had uniform backgrounds.} A. Correlation between human RDM and Krizhevsky et. al. 2012 RDM (Kendall $ \tau_{a} $ rank correlation), across different layers and levels of variations. Each color-coded curve shows the correlation of one layer of the model (specified on the right legend) with the corresponding human RDM. The correlation of Pixel representation with human RDM is depicted using a dashed, dark purple curve. The color-coded points on the top of the plots indicate whether the correlation is significant. Blue points indicate significant correlation while gray points show insignificant correlation. Correlation values are the average over 10,000 bootstrap resamples. Error bars are the standard deviation. B-H. Idem for Hybrid-CNN, Overfeat, CNN-F, CNN-M, CNN-S, Zeiler and Fergus, and HMAX, respectively. I. The average correlation across all levels for each layer of each model (error bars are STD). Each curve corresponds to one model. The shaded area shows the average correlation for the Pixel representation across all levels. All correlation values were calculated using the RSA toolbox (Nili et al., 2014).}
\label{fig11}
\end{figure*} 

\begin{figure*}[!htb]
\centering
\includegraphics[scale=0.8]{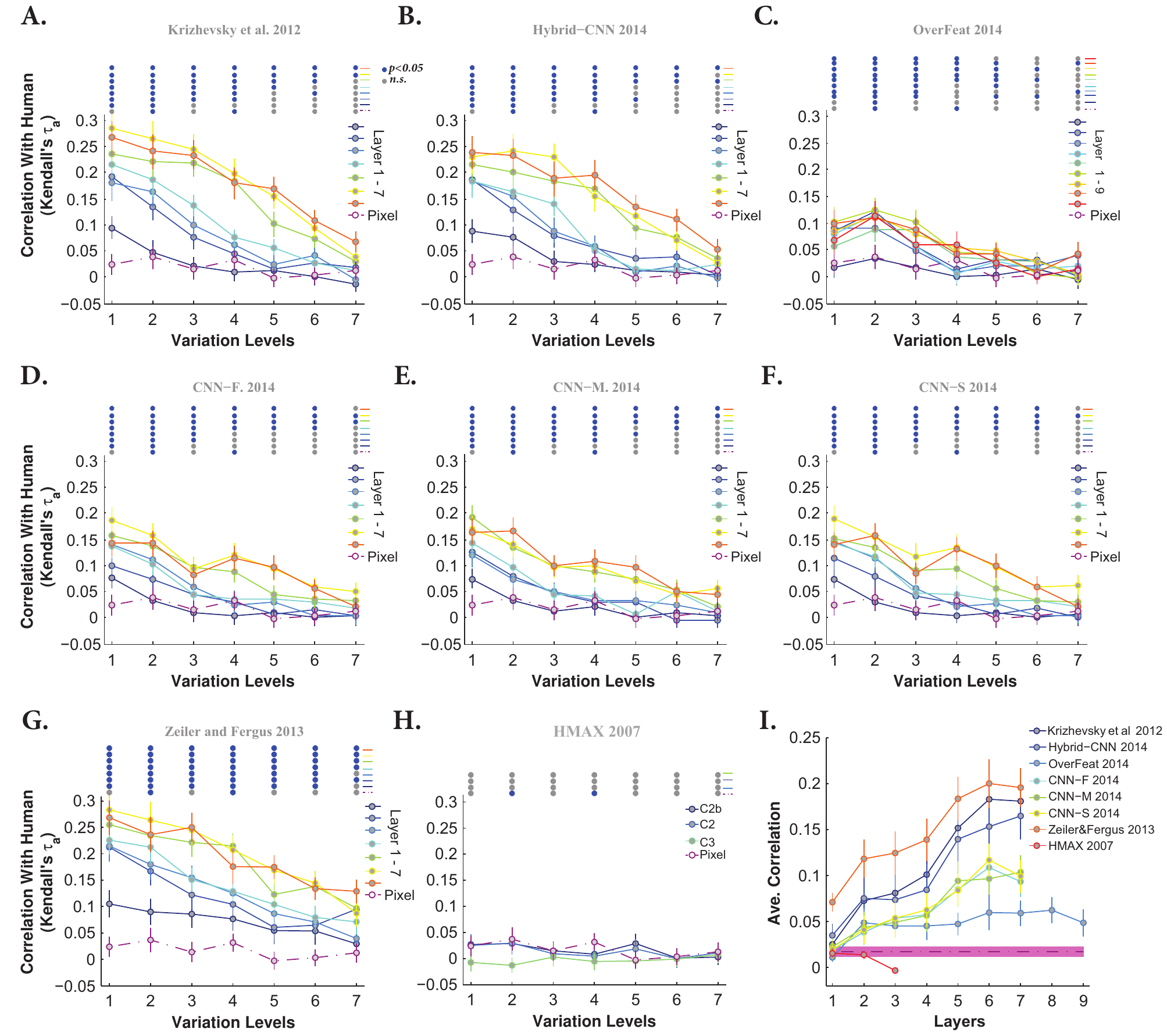}
\caption{{\bf Correlation between humans' and models' RDMs, across different layers and levels, when objects had natural backgrounds.} A-H. Correlation between humans' RDM and the one of KirZhevsky, Hybrid-CNN, Overfeat, CNN-F, CNN-M, CNN-S, Zeiler and Fergus, and HMAX, across all layers and levels. Figure conventions are identical to Fig.~\ref{fig11}. I. The average correlation across all levels for each layer of each model (error bars are STD).}
\label{fig12}
\end{figure*}

In contrast, DCNNs demonstrate clear categorical information for different objects across almost all levels, for both background conditions. Categorical information is more evident when objects had uniform background, even at high variation levels, while this information almost disappears at intermediate levels when object had natural backgrounds.  In addition, Overfeat did not clearly represent different object categories. The Overfeat model is one of the most powerful DCNNs with high accuracy on the Imagenet database, but it seems that the features are not suitable for our invariant object recognition task.  It uses no fewer than 230400 features! This might be one  reason for poor representational power: it probably leads to a nested and complex object representation. Besides, this high number of features may also explain the poor classification performance we obtained, due to overfitting.

Based on visual inspection, it seems that some DCNNs are better at representing some specific categories. For example, Krizhevsky, Hybrid-CNN, Zeiler and Fergus could better represent animal, car and airplane classes (lower within-class dissimilarity for these categories), while ship and motorcycle classes are better represented by CNN-F, CNN-M, and CNN-S. Interestingly, this has been reflected on the confusion matrix analysis, suggesting that combining and remixing of features from these DCNNs could result in a more robust invariant object representation~\cite{khaligh2014deep}. 

To quantify the similarity between models' and humans' RDMs, we calculated the correlation between them across all layers and levels (measured as Kendall $ \tau_{a} $ rank correlation).  Each panel in Fig.~\ref{fig11} and Fig.~\ref{fig12} represents the correlation between models' and humans' RDMs across all layers and variation levels  (each color-coded curve corresponds to one layer) when object had uniform and natural backgrounds, respectively. Overall, as  shown in these figures, the correlation coefficients are high at low variation levels , but decrease at higher levels. Moreover, correlations are not significant at very difficult levels, as specified with color-coded points on the top of each plot (blue point: significant, gray point: insignificant).

 Interestingly, comparing the cases of uniform (Fig.~\ref{fig11}) and natural (Fig.~\ref{fig12}) backgrounds indicates that the maximum correlation ($\thicksim 0.3$ at level 1) did not change a lot. However, for the uniform background condition, the correlation across other levels increased to some extent. Besides, it can also be seen that the correlations of the HMAX model and Pixel representation are higher and more significant than with natural backgrounds (Fig.~\ref{fig11}H and Fig.~\ref{fig12}H). 
 Note that the correlation values of the first layer  of almost all DCNNs (but Zeiler and Fergus) are similar to those of Pixel representation, suggesting that in the absence of viewpoint variations, very simple features (i.e., gray values of pixels) can achieve acceptable accuracy and correlation. This means that DCNNs are built to perform more complex recognition tasks, as it has been shown in several studies.

Not surprisingly, in the case of natural background, the correlation between Pixel and human RDMs are very low and almost insignificant at all levels (Fig.~\ref{fig12} dashed dark purple line copied on all panels). Similarly, the HMAX model shows a very low and insignificant correlation across all layers and levels.  We also expected a low correlation for the Overfeat model, as shown in Fig.~\ref{fig12}C. Interestingly, the correlation increases as images are processed across consecutive layers in DCNNs, with lower correlations at early layers and higher correlations at top layers (layer 5, 6, and 7). As for the accuracy results, the correlations of fully connected  layers of DCNNs are very similar to each other, suggesting that these layers do not greatly add to the final representation. 

We summarized the correlation results in Fig.~\ref{fig11}I and Fig.~\ref{fig12}I, by averaging the correlation coefficients across levels for every model layer.  It is shown that the correlations for DCNNs  evolve across layers, with low correlations at early layers and high correlations at top layers. Moreover, Fig.~\ref{fig11}I shows that the correlation of the HMAX model (all the layers) with human fluctuates around the correlation of Pixel representation (specified with shaded area).

Note that although the correlation coefficients are not very high ($\thicksim 0.2$), Zeiler and Fergus, Hybrid-CNN, and Krizhevsky models are the most human-like. It is worth noting that the best models in terms of performance, CNN-F, CNN-M, and CNN-S do not have the most human-like RDMs. Conversely, the model with the most human-like RDM, Zeiler and Fergus, is not the best in terms of classification performance.

More research is needed to understand why the Zeiler and Fergus' RDM is significantly more human-like than those of other DCNNs. This finding is consistent with a previous study by Cadieu et al.\cite{cadieu2014deep}, in which the Zeiler and Fergus' RDM was found be more similar to monkey IT RDM than those of the Krizhevsky and HMAX models.

We also computed the category separability index for the internal representations of each model by computing the ratio of within-category relative to between-category dissimilarities (results are not shown here).
This experiment also confirms that models with higher separability indexes do not necessarily perform better than other models. In fact, it is the actual positions of images of different categories in the representational space which determines the final accuracy of a model, not just the mean inter- and intra-class distances. 
\subsection*{A very deep network}
In previous sections we studied different DCNNs, each having 8 or 9 layers with 5 or 6 convolutional layers, from various perspectives and compared them with the human feed-forward object recognition system. Here, we assess how exploiting many more layers could affect the performance of DCNNs. To this end, we used Very Deep CNN~\cite{simonyan2014very} that has no fewer than 19 layers (16 convolutional and 3 fully connected layers). We extracted features of layers 9 to 18 from images with natural backgrounds, to investigate if more layers in the Very Deep CNN affects the final accuracy and human-likeness.

\begin{figure*}[!htb]
\centering
\includegraphics[scale=0.8]{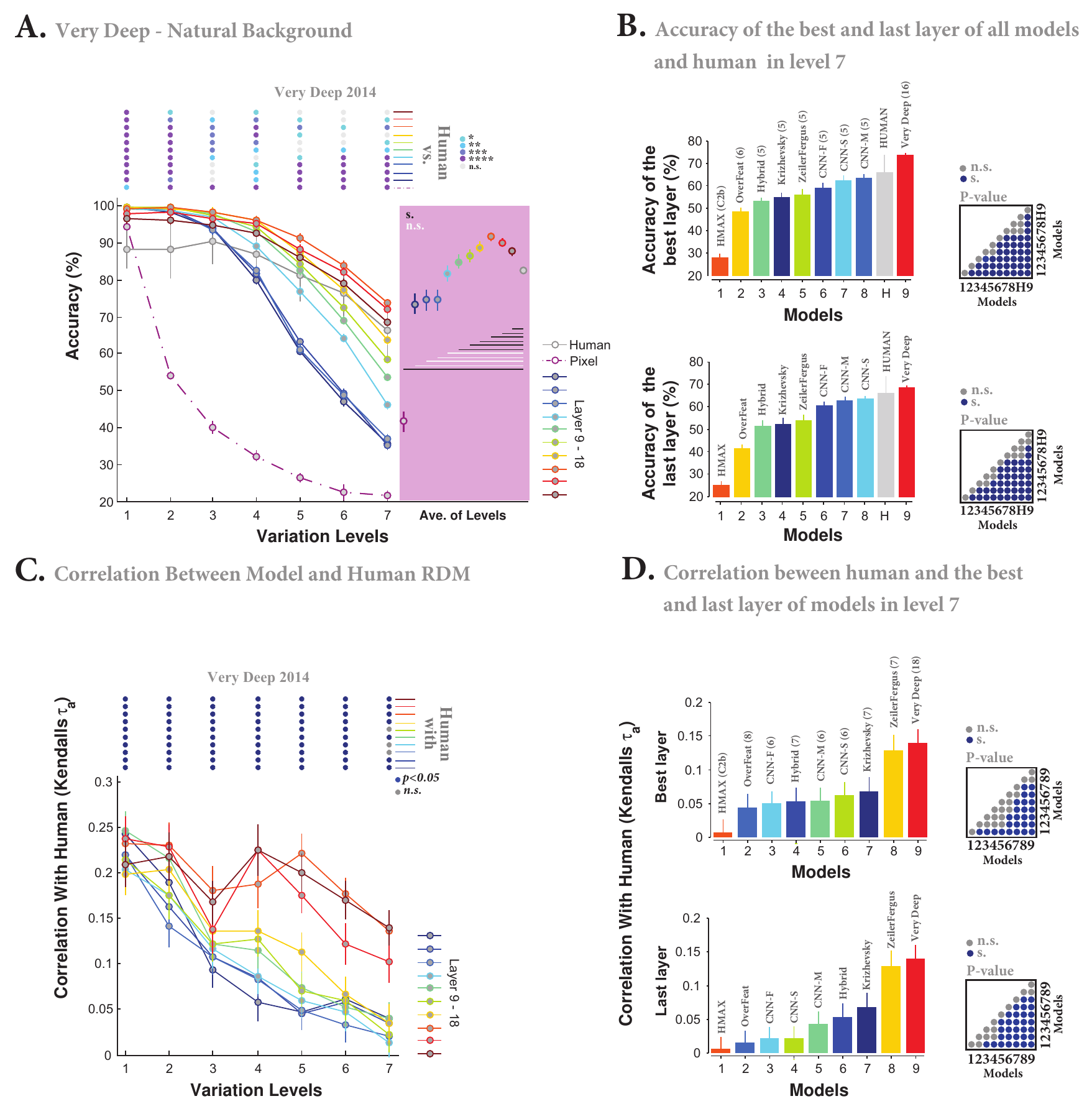}
\caption{{\bf The accuracy and human-likeness of the Very Deep CNN with natural backgrounds.} A. Classification accuracy of the Very Deep CNN (layers 9-18 ) and humans across the seven levels of object variations. Each colored curve shows the accuracy of one layer of the model. The accuracy of the Pixel representation is depicted using a dashed, dark purple curve. The gray curve indicates human categorization accuracy across the seven levels. The color-coded points at the top of the plot indicate whether there is a significant difference between the accuracy of humans and each layer of the model (Wilcoxon rank sum test). Each color refers to a p-value, specified on the top-right ($*$: $p<0.05$, $**$: $p<0.01$, $***$: $p<0.001$, $****$: $p<0.0001$). We plot the mean accuracies +/- STD over 15 runs. Colored circles with error bars, on the pink area show the average accuracy of each layer across all variation levels (mean +/- STD). The horizontal lines underneath the circles, indicate whether the difference between human accuracy (gray circle) and each layer of the model is significant (Wilcoxon rank sum test; black line: significant, white line: insignificant). B. Top: the accuracy comparison between the best-performing layer in each model and humans at the last variation level (level 7). The color-coded matrix, on the right of the bar plot, shows the p-values for all pairwise comparisons between humans and models (Wilcoxon rank sum test). Numbers, written around the p-value matrix, correspond to models (H stands for human). Bottom: idem with the last layers. C. Correlation between humans and the Very Deep CNN RDMs, across different layers (layers 9-18) and levels. Each color-coded curve shows the correlation of one layer of the model  with corresponding human RDM. The color-coded points at the top of the plot indicate whether the correlation is significant (Blue: significant; Gray: insignificant). Correlation values are the average over 10,000 bootstrap resamples +/- STD. D. Top: correlations between the most correlated layer in each model and humans at the last variation level (level 7). P-value matrix was calculated using similar approach to B. Bottom: idem with the last layers.}
\label{fig13}
\end{figure*}

Figure~\ref{fig13}A illustrates that the classification accuracy tends to improve as images are processed through consecutive layers. The accuracies of layers 9, 10, and 11 are almost the same. But, the accuracy gradually increases over the next layers and culminates in layer 16 (the topmost convolutional layer), which significantly outperforms humans even at the highest variation level (see the color-coded circles above this figure). Here again, the accuracy drops in fully connected layers that are optimized for the Imagenet classification. Nevertheless, the accuracies of the highest layer (layer 18) are still higher than those of humans for all variation levels.

Figure~\ref{fig13}B demonstrates the accuracies of the last and best-performing layers of all models in comparison with humans for the highest variation level (level 7) in the natural background task. The color-coded matrix on the right shows the p-values for all pairwise comparisons between models and humans computed by the Wilcoxon rank sum test. It can be seen that the Very Deep CNN significantly outperforms all other DCNNs in both cases. It is also evident that the best-performing layer of this model significantly outperforms humans. However, the accuracies of all other DCNNs are below the humans, and the gap is significant for all models  but CNN-S and CNN-M.

\begin{figure}[!htb]
\centering
\includegraphics[scale=0.5]{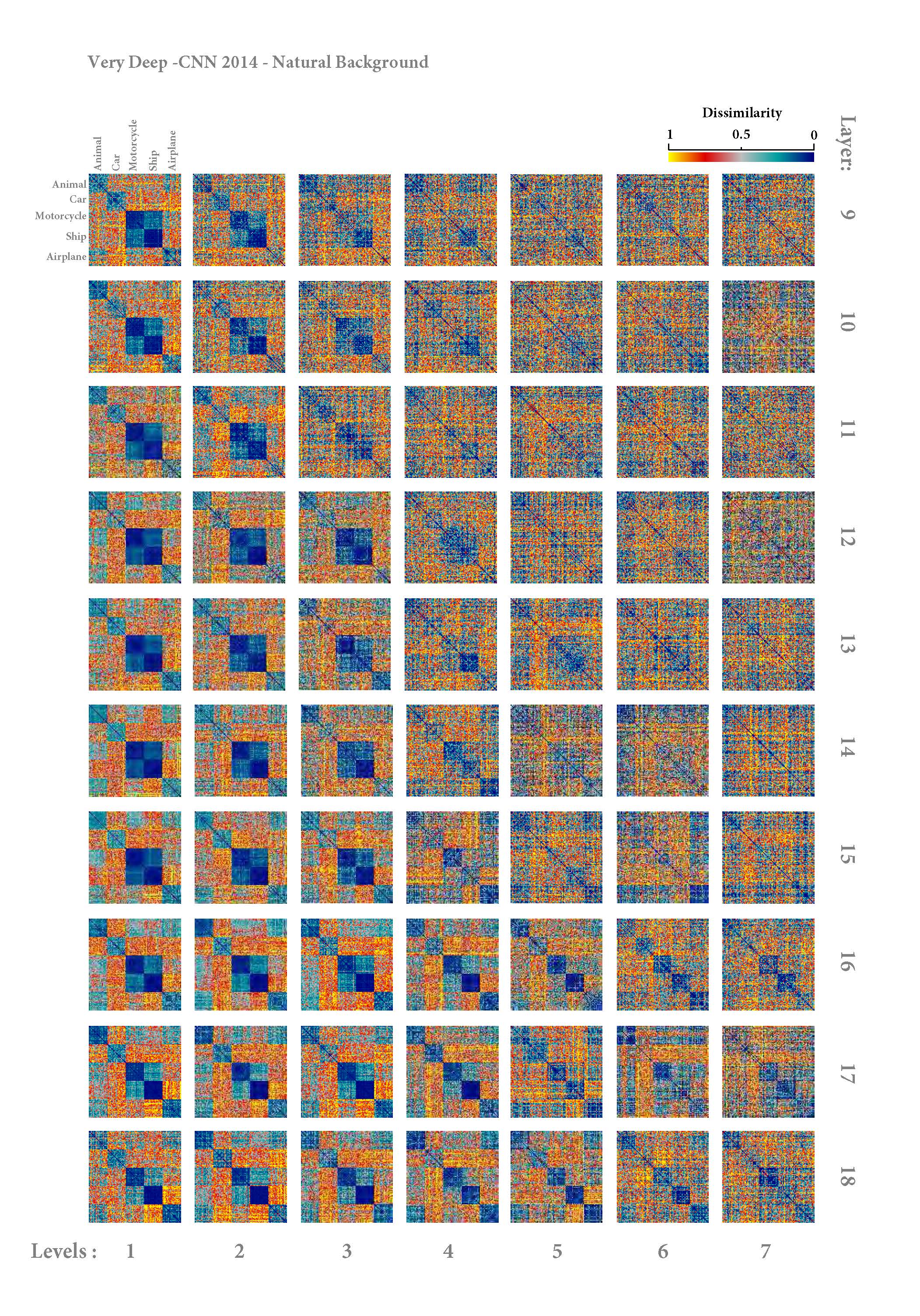}
\caption{Representational Dissimilarity Matrices (RDM) of Very Deep  model (layers 9 to 18) for different levels of variation (from level 1-7) in  natural background condition. Each element in a matrix shows the pairwise dissimilarities between the representations  of two images (measured as 1-r, Spearman's rank correlation. See Materials and Methods). The color bar at the top-right corner shows the degree of dissimilarity. The size of each matrix is $100\times 100$, with 20 images from each category. This was done for the sake of better visualization.}
\label{fig14}
\end{figure}

We also computed the RDM of the Very Deep model for all variation levels and layers 9 to 18 in the natural background condition (see Fig.~\ref{fig14}). Calculating the correlations between the model's and humans' RDMs shows that the last three  layers had the highest correlations with human RDM (see Fig.~\ref{fig13}C). The correlation values of other layers drastically decrease down to 0.05, indicating that these layers are less robust to object variations than the last layers. However, the statistical analysis demonstrates that almost all correlation values are significant (see color-coded points above the plot), suggesting that although the amount of similarity between the RDM of humans and that of the Very Deep model's layers are small, these similarities are not random but statistically meaningful. Hence, it can be said that the layers of Very Deep CNN process images in a somewhat human-like way. Finally, Fig.~\ref{fig13}D compares the correlation values between the RDM of humans and the one of the last as well as the best-correlated layers of all DCNNs in the natural background condition. As can be seen, the Very Deep CNN and Zeiler and Fergus models have the highest correlation values in both cases, with large statistical difference compared to other models. 

\section*{Discussions}

Invariant object recognition has always been a demanding task to solve in computer vision, yet it is simply done by a two-year old child. However, the emergence of novel learning mechanisms and computational models in recent years has opened new avenues for solving this highly complex task. DCNNs have been shown to be a novel and powerful approach to tackle this problem~\cite{Krizhevsky2012,dosovitskiy2014discriminative,jones2014computer,zeiler2014visualizing,girshick2014rich, alemi2013multifeatural,szegedy2014going,cadieu2014deep,le2013building,mohamed2011deep}. These networks have drawn scientists' attention not only in vision sciences, but also in other fields of science (see~\cite{jones2014computer}), as a powerful solution for many complex problems. DCNNs are among the most powerful computing models inspired by computations performed in neural circuits. To our interest, recent studies also confirmed the abilities of DCNNs in object recognition problems (e.g.~\cite{Krizhevsky2012},~\cite{cadieu2014deep}, and~\cite{donahue2013decaf}). Besides, several studies have tried to compare the responses of DCNNs and primate visual cortex in different object recognition tasks.

Khaligh-Razavi and Kriegeskorte~\cite{khaligh2014deep} compared the representational geometry of neuronal responses in human (fMRI data; see~\cite{kriegeskorte2008matching}) and monkey  IT cortex (cell recording; see~\cite{kiani2007object}) with several computational models, including one DCNN, on a 96-image dataset. They showed that supervised DCNNs can explain IT representation. However, firstly, their image database only contained frontal views of objects with no viewpoint variation. Secondly, the number and variety of images were very low (only 96 images), compared to the wide variety of complex images in natural environment. Finally, images had a uniform gray background, which is very different from natural vision. To overcome such issues, Cadieu et. al.~\cite{cadieu2014deep} used a large image database, consisting of different categories, backgrounds, transformations, and compared the categorization accuracy and representational geometry of three DCNNs and neural responses in IT and V4 of monkey. They showed that DCNNs closely resemble the responses of IT neurons either in accuracy or geometry~\cite{cadieu2014deep,yamins2014performance}. One issue in their study is the long stimulus presentation time (100 ms), which might be too long to only account for feed-forward processing. Moreover, they included only three DCNNs in their study. In another attempt, G{\"u}{\c{c}}l{\"u} et. al.~\cite{gucclu2014deep} mapped different layers of a DCNN onto the human visual cortex.  More specifically, they computed the representational similarities among different layers of a DCNN and the fMRI data from different areas in human visual cortex. Although these studies have shown the power of several DCNNs in object recognition, advancements in developing new DCNNs are quick, which requires continuous assessments of recent DCNNs using different techniques. Moreover, the ability of DCNNs in tolerating object variations (mostly 3-D variations) had not been carefully evaluated before. 

Here, we comprehensively tested eight best performing DCNNs, reported in the literature~\cite{Krizhevsky2012,zeiler2014visualizing,sermanet2013overfeat, chatfield2014return,zhou2014learning,simonyan2014very}, in a very challenging vision task, namely invariant object recognition. This list of DCNNs has shown remarkable accuracies in classification of big and challenging image databases such as Imagenet, VOC 2007, and Caltech 205. Moreover, we compared the DCNNs with human subjects performing the same task with the same images to investigate the extent to which DCNNs resemble humans. 

\subsection*{DCNNs achieve human-level performance in rapid invariant object recognition task}
Humans are very fast and accurate at categorizing  objects ~\cite{thorpe1996speed,vanrullen2001time,fabre2011characteristics}. Numerous studies have investigated this remarkable performance  under ultra-rapid image presentation~\cite{kirchner2006ultra,mack2011timing,potter2014detecting}.  It is believed that rapid object categorization is mainly performed by the feed-forward information flow through the ventral visual pathway~\cite{kreiman2007limits,fabre2011characteristics}. Experimental and theoretical evidence suggests that feed-forward processing is able to perform invariant object recognition~\cite{liu2009timing,freiwald2010functional,anselmi2014unsupervised,hung2005fast}. Here, we measured human accuracy when categorizing five object categories in a rapid presentation paradigm. Objects varied in six dimensions and the task difficulty was controlled using seven variation levels. Results showed that humans achieved high accuracy across all levels (under 2- and 3-D variations) while objects were only presented for 25 ms. 

Using the same image database, we also evaluated eight state-of-the-art DCNNs~\cite{Krizhevsky2012,zeiler2014visualizing,sermanet2013overfeat,chatfield2014return,zhou2014learning}, largely inspired by feed-forward processing of visual cortex. Results indicated that these DCNNs can mimic human accuracy (see Fig.~\ref{fig2} to Fig.~\ref{fig5}). However, the HMAX model, as one of the early successful models, showed very poor performance in almost all experiments. We also showed in our previous study that such shallow feed-forward models fails to achieve human-level accuracy in invariant object categorization~\cite{ghodrati2014feedforward}.

We further performed layer-specific analysis to investigate how accuracy and representational geometry evolve across consecutive layers in DCNNs. Results illustrated that accuracies tend to increase as images are processed through the layers; however, some layers achieved very similar accuracies. If some layers do not considerably contribute to the final accuracy, at least in our task, one is tempted to remove it, to reduce the computational load of the DCNN, which is typically very high. For example it has been shown that eliminating one of the middle layers leads to just 2\% accuracy drop in Krizhevsky model on the Imagenet database~\cite{Krizhevsky2012}. More research is needed to systematically evaluate the role of different layers by removing each layer and evaluating the resulting accuracy. However, this should be done using different image databases since these DCNNs were optimized for Imagenet database. Therefore, the layer-specific effect might be database dependent.

The layer-specific analysis is interesting as it shows that not only the accuracy, but also the representational geometry evolves through layers. To our knowledge, only one study~\cite{khaligh2014deep} had investigated the layer-specific responses in one DCNN. A possible future study would be comparing the responses of several visual cortical areas with different layers of DCNNs as it helps to understand what is missing in models and layers. Cadieu et.~al.~\cite{cadieu2014deep} compared the responses of monkey IT and V4 neurons with the penultimate layer of three DCNNs, but they did not tested, for example, how V4 responses are correlated to other layers.

RDMs (Fig.~\ref{fig10}) and confusion matrices (Fig.~\ref{fig6}) of the last layer of DCNNs demonstrated that increasing the level of object variations can disturb object representations and  increase the misclassification rate, but less so for the higher layers. Conversely, for low variation levels, shallow models actually outperform both deeper ones and humans. This means that, even if deep nets have attracted a lot of attention recently, deeper is not always better. To classify images with weak viewpoint variations (e.g. passport photos), a shallow model might lead to the best performance. In addition, its computational load will be much lower, and training will require much fewer labeled examples.


It is possible, and even likely, that having incongruent backgrounds can affect the human accuracy in some cases. However, we ran the same exact experiments with uniform backgrounds. This helped us to find  an upper bound for the human performance (see Fig.~\ref{fig2}). Even in this case, models can reach human-level accuracy.  Moreover, since both humans and DCNNs saw the objects in a congruent context during the development, eliminating the contextual information in the background, or using an incongruent background, presumably similarly affect the humans and the models.

In summary, our results demonstrate the ability of DCNNs to reach human (feed-forward vision) accuracy in invariant object recognition. This confirms the success of these computational models to mimic the performance of the visual neural circuits in such a difficult task. When variation level is high, shallow networks have low accuracies, while as we move through the layers of DCNNs the invariance gradually increase in such a way that the Very Deep network (with 19 layers) can even outperform the humans. Another important point is that both 2-D and 3-D variations could be handled by 2-D features extracted through the layers of DCNNs. Although some 2-D variations, such as position, are treated through many convolutional layers (using shared-weight filters in different positions), DCNNs do not have any built-in mechanism to overcome 3-D variations (such as in-depth-rotation). Thus, these invariances must be learned. Regarding to different theories of how the brain reaches 3-D invariance, our results suggest that 3-D rotation invariance can be achieved using 2-D features and not necessarily by construction of 3-D object models. However, the difference between the error distributions and object representations of DCNNs and humans suggest that they use different information to handle invariant object recognition, presumably due to structural and learning differences. The human visual system exploits feedback signals, bottom-up and top-down attentions, continuous visual information, and temporal learning. So if using more layers can substantially improve the performance of machine vision algorithms, adding other properties of the visual system can make more advances. This could, in reverse, give important clues about the nature of neural processing in the visual cortex.
\subsection*{Network architecture plays a very important role}
Here, we evaluated several DCNNs with different architectures and  training sets, which led to different accuracies. Zeiler and Fergus, CNN-M and CNN-S achieved higher accuracies than Keizhevsky model, while they used smaller receptive fields and smaller stride in the first convolutional layer. Besides, CNN-M and CNN-S outperformed Zeiler and Fergus using more convolutional features in layers 3, 4 and 5. Nevertheless, Overfeat that exploits extensively more features in these layers had troubles with invariant object recognition. Interestingly, Very Deep CNN,  which significantly outperforms all models as well as humans,  has about twice convolutional layers as other DCNNs but smaller ($3 \times 3$) receptive fields. 

Although it is not clear why some DCNNs perform better than others, our results suggest that networks with deeper architecture, and convolutional layers with small filter size but with more feature planes can achieve higher performances. In any case, an extensive optimization is required to find the best architecture and parameter settings for DCNNs. It is also important to point out that despite utilizing similar architectures but different training datasets, Keizhevsky and Hybrid-CNN models had close performances. These results suggest that architecture is more important than the training set. Hence, future studies should focus on how to evaluate different architectures to find the optimum one.

\subsection*{DCNNs lack important processing mechanisms that exist in biological vision}

We tried to only allow  feed-forward processing in our psychophysical experiment by using short presentation time and backward masking, weakening the effect of back projections. However, this does not completely rule out the effects of feedback connections in the visual system. Conversely, DCNNs are feed-forward only models without any feedback mechanisms from upper to lower layers (note that error back propagation is not considered as a feedback mechanism because it only occurs during the learning, not the recognition). Adding a feedback mechanism to DCNNs could increase their performance, and this could be useful for complex visual tasks (e.g., variation level 7 in our data). This would inevitably increases the computational load of DCNNs and that might be the reason why DCNNs still lack a feedback mechanism. Another issue is how to learn feedback connections. 

In addition to object recognition, feedback connections plays a pivotal role in other visual processes such as figure-ground segregation~\cite{roelfsema2002figure,raudies2010neural}, spatial and feature-based attention~\cite{gilbert2013top}, and perceptual learning~\cite{pannunzi2012learning}. As shown in our results, the accuracies of DCNNs significantly drops in case of objects with natural backgrounds. This could be due to the lack of a figure-ground segregation in the models. Indeed, the primate visual system is able to separate the parts of image which belong to the target object from the background and other objects. It has been suggested that recurrent processing is required for the completion of figure-ground segregation (see ~\cite{roelfsema2002figure} and~\cite{raudies2010neural}). Also, the mechanisms of bottom-up and top-down attention in the human visual system emphasizes the most salient and relevant parts of the images, which contain more information and can facilitate the categorization process. Several studies~\cite{lamme2000distinct,wyatte2012limits,o2013recurrent} have shown that recurrent processing can enhance object representations in IT and facilitate invariant object recognition. DCNNs lack such mechanisms, and they could help to increase the recognition accuracy, especially in cluttered images and this could be another direction for future improvement of DCNNs.
\subsection*{Future directions}
Our image database has several advantages for studying  invariant object recognition. Firstly, it contains a large number of object images, changing across different levels of variations of position, scale, and in-depth and in-plane rotations, and background. Secondly, we had a precise control over the amount of variations  that let us generate images with different degrees of complexity/difficulty; Therefore, enabling us to scrutinize the behavior of humans and computational models, while the complexity of object variations gradually increases. Thirdly, similar to several studies~\cite{yamins2014performance,cadieu2014deep,rajalingham2015comparison,majaj2015simple}, by eliminating dependencies between objects and backgrounds, we were able to study invariance, independently of contextual effects.

However, there are several effective parameters in invariant object recognition for both humans and models that should be further investigated. It is important to explore how the consistency between objects and surrounding environment would affect the object recognition process~\cite{sastyin2015does,oliva2007role,joubert2008early,remy2013object} and it should be further studied in invariant object recognition. Also, other parameters such as illumination, contrast, texture, noise, and occlusion need to be investigated in controlled experiments. 

Another important question that needs to be clearly addressed is whether all types of variations impose the same difficulty to humans and models. A simple and short answer is ``No"; however, it remains unclear which types of variation are more challenging, what  the underlying mechanisms for it are. It has been shown that the brain responds differently to different types of object variations. For instance, scale invariant responses appear faster than position invariant ones~\cite{isik2014dynamics}. Interestingly, scale invariant responses in the human brain emerge early in development while view invariance responses tend to emerge later, suggesting that simple processes such as scale invariance could be built-in, while we would need more training to perform view invariant object recognition~\cite{nishimura2014size}.  Therefore, it is important, for both neuroscientists and computational modelers, to understand how the brain deals with different types of variations. From a computer vision point of view, it seems that 3-D variations (e.g., rotations in-depth) are more challenging than 2-D transformations (e.g., changes in position and scale)~\cite{Pinto2008,pinto2011comparing,yamins2014performance}. Due to the structure of DCNNs and the computations performed in such networks, they easily tackle with changes in position and, to some extent, in the scale of the objects. However, there is no built-in mechanism for invariance to 3-D transformations.  Adding such a mechanism to the models should increase their accuracy as well as their resemblance to neurophysiological data. A very recent modeling study~\cite{farzmahdi2015specialized}, inspired by physiological data from monkeys brain, shows that adding a view invariance mechanism to a feed-forward model can surprisingly explain face processing in monkey face patches~\cite{tsao2003faces,tsao2006cortical}. 

\footnotesize
\bibliographystyle{elsarticle-num}

\begin{thebibliography}{10}
\expandafter\ifx\csname url\endcsname\relax
  \def\url#1{\texttt{#1}}\fi
\expandafter\ifx\csname urlprefix\endcsname\relax\def\urlprefix{URL }\fi
\expandafter\ifx\csname href\endcsname\relax
  \def\href#1#2{#2} \def\path#1{#1}\fi

\bibitem{dicarlo2012does}
J.~J. DiCarlo, D.~Zoccolan, N.~C. Rust, How does the brain solve visual object
  recognition?, Neuron 73~(3) (2012) 415--434.

\bibitem{dicarlo2007untangling}
J.~J. DiCarlo, D.~D. Cox, Untangling invariant object recognition, Trends in
  Cognitive Sciences 11~(8) (2007) 333--341.

\bibitem{liu2009timing}
H.~Liu, Y.~Agam, J.~R. Madsen, G.~Kreiman, Timing, timing, timing: fast
  decoding of object information from intracranial field potentials in human
  visual cortex, Neuron 62~(2) (2009) 281--290.

\bibitem{freiwald2010functional}
W.~A. Freiwald, D.~Y. Tsao, Functional compartmentalization and viewpoint
  generalization within the macaque face-processing system, Science 330~(6005)
  (2010) 845--851.

\bibitem{thorpe1996speed}
S.~Thorpe, D.~Fize, C.~Marlot, et~al., Speed of processing in the human visual
  system, Nature 381~(6582) (1996) 520--522.

\bibitem{anselmi2014unsupervised}
F.~Anselmi, J.~Z. Leibo, L.~Rosasco, J.~Mutch, A.~Tacchetti, T.~Poggio,
  Unsupervised learning of invariant representations with low sample
  complexity: the magic of sensory cortex or a new framework for machine
  learning?, arXiv:1311.4158.

\bibitem{hung2005fast}
C.~P. Hung, G.~Kreiman, T.~Poggio, J.~J. DiCarlo, Fast readout of object
  identity from macaque inferior temporal cortex, Science 310~(5749) (2005)
  863--866.

\bibitem{Fukushima1980}
K.~Fukushima, Neocognitron : a self organizing neural network model for a
  mechanism of pattern recognition unaffected by shift in position., Biological
  Cybernetics 36~(4) (1980) 193--202.

\bibitem{LeCun1998}
Y.~LeCun, Y.~Bengio, Convolutional networks for images, speech, and time
  series, in: The Handbook of Brain Theory and Neural Networks, MIT Press,
  1998, pp. 255--258.

\bibitem{Serre2007.PAMI}
T.~Serre, L.~Wolf, S.~Bileschi, M.~Riesenhuber, T.~Poggio, Robust object
  recognition with cortex-like mechanisms, IEEE Transactions on Pattern
  Analysis Machine Intelligence 29~(3) (2007) 411--426.

\bibitem{Masquelier2007}
T.~Masquelier, S.~J. Thorpe, Unsupervised learning of visual features through
  spike timing dependent plasticity., PLoS Computational Biology 3~(2) (2007)
  e31.

\bibitem{Lee2009}
H.~Lee, R.~Grosse, R.~Ranganath, A.~Y. Ng, Convolutional deep belief networks
  for scalable unsupervised learning of hierarchical representations,
  Proceedings of the 26th Annual International Conference on Machine Learning
  (2009) 1--8.

\bibitem{cox2014neural}
D.~D. Cox, T.~Dean, Neural networks and neuroscience-inspired computer vision,
  Current Biology 24~(18) (2014) R921--R929.

\bibitem{schmidhuber2015deep}
J.~Schmidhuber, Deep learning in neural networks: An overview, Neural Networks
  61~(1) (2015) 85--117.

\bibitem{Krizhevsky2012}
A.~Krizhevsky, I.~Sutskever, G.~Hinton, Imagenet classification with deep
  convolutional neural networks., in: Neural Information Processing Systems
  (NIPS), 2012, pp. 1--9.

\bibitem{zeiler2014visualizing}
M.~D. Zeiler, R.~Fergus, Visualizing and understanding convolutional networks,
  in: European Conference on Computer Vision, 2014, pp. 818--833.

\bibitem{sermanet2013overfeat}
P.~Sermanet, D.~Eigen, X.~Zhang, M.~Mathieu, R.~Fergus, Y.~LeCun, Overfeat:
  Integrated recognition, localization and detection using convolutional
  networks, arXiv:1312.6229.

\bibitem{chatfield2014return}
K.~Chatfield, K.~Simonyan, A.~Vedaldi, A.~Zisserman, Return of the devil in the
  details: Delving deep into convolutional nets, arXiv:1405.3531.

\bibitem{ghodrati2014feedforward}
M.~Ghodrati, A.~Farzmahdi, K.~Rajaei, R.~Ebrahimpour, S.-M. Khaligh-Razavi,
  Feedforward object-vision models only tolerate small image variations
  compared to human, Frontiers in Computational Neuroscience 8~(74) (2014)
  1--17.

\bibitem{khaligh2014deep}
S.-M. Khaligh-Razavi, N.~Kriegeskorte, Deep supervised, but not unsupervised,
  models may explain it cortical representation, PLoS Computational Biology
  10~(11) (2014) e1003915.

\bibitem{pinto2011comparing}
N.~Pinto, Y.~Barhomi, D.~D. Cox, J.~J. DiCarlo, Comparing state-of-the-art
  visual features on invariant object recognition tasks, in: IEEE workshop on
  Applications of Computer Vision, 2011, pp. 463--470.

\bibitem{Pinto2008}
N.~Pinto, D.~D. Cox, J.~J. DiCarlo, Why is real-world visual object recognition
  hard?, PLoS Computational Biology 4~(1) (2008) e27.

\bibitem{Liu2015}
J.~Liu, B.~Liu, H.~Lu, Detection guided deconvolutional network for
  hierarchical feature learning, Pattern Recognition 48~(8) (2015) 2645--2655.

\bibitem{yosinski2014transferable}
J.~Yosinski, J.~Clune, Y.~Bengio, H.~Lipson, How transferable are features in
  deep neural networks?, in: Advances in Neural Information Processing Systems,
  2014, pp. 3320--3328.

\bibitem{peng2014exploring}
X.~Peng, B.~Sun, K.~Ali, K.~Saenko, Exploring invariances in deep convolutional
  neural networks using synthetic images, arXiv:1412.7122.

\bibitem{cheung2014discovering}
B.~Cheung, J.~A. Livezey, A.~K. Bansal, B.~A. Olshausen, Discovering hidden
  factors of variation in deep networks, arXiv:1412.6583.

\bibitem{cadieu2014deep}
C.~F. Cadieu, H.~Hong, D.~L. Yamins, N.~Pinto, D.~Ardila, E.~A. Solomon, N.~J.
  Majaj, J.~J. DiCarlo, Deep neural networks rival the representation of
  primate it cortex for core visual object recognition, PLoS Computational
  Biology 10~(12) (2014) e1003963.

\bibitem{gucclu2014deep}
U.~G{\"u}{\c{c}}l{\"u}, M.~A. van Gerven, Deep neural networks reveal a
  gradient in the complexity of neural representations across the brain's
  ventral visual pathway, arXiv:1411.6422.

\bibitem{lecun2015deep}
Y.~LeCun, Y.~Bengio, G.~Hinton, Deep learning, Nature 521~(7553) (2015)
  436--444.

\bibitem{jia2014caffe}
Y.~Jia, E.~Shelhamer, J.~Donahue, S.~Karayev, J.~Long, R.~Girshick,
  S.~Guadarrama, T.~Darrell, Caffe: Convolutional architecture for fast feature
  embedding, in: Proceedings of the ACM International Conference on Multimedia,
  2014, pp. 675--678.

\bibitem{zhou2014learning}
B.~Zhou, A.~Lapedriza, J.~Xiao, A.~Torralba, A.~Oliva, Learning deep features
  for scene recognition using places database, in: Advances in Neural
  Information Processing Systems, 2014, pp. 487--495.

\bibitem{simonyan2014very}
K.~Simonyan, A.~Zisserman, Very deep convolutional networks for large-scale
  image recognition, arXiv:1409.1556.

\bibitem{serre2007feedforward}
T.~Serre, A.~Oliva, T.~Poggio, A feedforward architecture accounts for rapid
  categorization, Proceedings of the National Academy of Sciences 104~(15)
  (2007) 6424--6429.

\bibitem{hubel1962receptive}
D.~H. Hubel, T.~N. Wiesel, Receptive fields, binocular interaction and
  functional architecture in the cat's visual cortex, The Journal of Physiology
  160~(1) (1962) 106--154.

\bibitem{hubel1968receptive}
D.~H. Hubel, T.~N. Wiesel, Receptive fields and functional architecture of
  monkey striate cortex, The Journal of Physiology 195~(1) (1968) 215--243.

\bibitem{MUTCH10}
J.~Mutch, U.~Knoblich, T.~Poggio, {CNS}: a {GPU}-based framework for simulating
  cortically-organized networks, Tech. Rep. MIT-CSAIL-TR-2010-013 / CBCL-286,
  Massachusetts Institute of Technology, Cambridge, MA (February 2010).

\bibitem{cauchoix2016fast}
M.~Cauchoix, S.~M. Crouzet, D.~Fize, T.~Serre, Fast ventral stream neural
  activity enables rapid visual categorization, NeuroImage 125 (2016) 280--290.

\bibitem{crouzet2011visual}
S.~M. Crouzet, T.~Serre, What are the visual features underlying rapid object
  recognition?, Frontiers in psychology 2 (2011) 326.

\bibitem{lamme2002masking}
V.~Lamme, K.~Zipser, H.~Spekreijse, et~al., Masking interrupts figure-ground
  signals in v1, Journal of Cognitive Neuroscience 14~(7) (2002) 1044--1053.

\bibitem{brainard1997psychophysics}
D.~H. Brainard, The psychophysics toolbox, Spatial Vision 10~(4) (1997)
  433--436.

\bibitem{breitmeyer2006visual}
B.~Breitmeyer, H.~{\"O}{\u{g}}men, Visual masking: Time slices through
  conscious and unconscious vision, Vol.~41, Oxford University Press, 2006.

\bibitem{lamme2000distinct}
V.~A. Lamme, P.~R. Roelfsema, The distinct modes of vision offered by
  feedforward and recurrent processing, Trends in Neurosciences 23~(11) (2000)
  571--579.

\bibitem{CC01a}
C.-C. Chang, C.-J. Lin, {LIBSVM}: A library for support vector machines, ACM
  Transactions on Intelligent Systems and Technology 2 (2011) 27:1--27:27.

\bibitem{nili2014toolbox}
H.~Nili, C.~Wingfield, A.~Walther, L.~Su, W.~Marslen-Wilson, N.~Kriegeskorte, A
  toolbox for representational similarity analysis, PLoS Computational Biology
  10~(4) (2014) e1003553.

\bibitem{kheradpisheh2015bio}
S.~R. Kheradpisheh, M.~Ganjtabesh, T.~Masquelier, Bio-inspired unsupervised
  learning of visual features leads to robust invariant object recognition,
  arXiv:1504.03871.

\bibitem{ghodrati2012can}
M.~Ghodrati, S.-M. Khaligh-Razavi, R.~Ebrahimpour, K.~Rajaei, M.~Pooyan, How
  can selection of biologically inspired features improve the performance of a
  robust object recognition model?, PloS one 7~(2) (2012) e32357.

\bibitem{yamins2014performance}
D.~L. Yamins, H.~Hong, C.~F. Cadieu, E.~A. Solomon, D.~Seibert, J.~J. DiCarlo,
  Performance-optimized hierarchical models predict neural responses in higher
  visual cortex, Proceedings of the National Academy of Sciences 111~(23)
  (2014) 8619--8624.

\bibitem{kriegeskorte2008matching}
N.~Kriegeskorte, M.~Mur, D.~A. Ruff, R.~Kiani, J.~Bodurka, H.~Esteky,
  K.~Tanaka, P.~A. Bandettini, Matching categorical object representations in
  inferior temporal cortex of man and monkey, Neuron 60~(6) (2008) 1126--1141.

\bibitem{kiani2007object}
R.~Kiani, H.~Esteky, K.~Mirpour, K.~Tanaka, Object category structure in
  response patterns of neuronal population in monkey inferior temporal cortex,
  Journal of Neurophysiology 97~(6) (2007) 4296--4309.

\bibitem{carlson2014reaction}
T.~A. Carlson, J.~B. Ritchie, N.~Kriegeskorte, S.~Durvasula, J.~Ma, Reaction
  time for object categorization is predicted by representational distance,
  Journal of Cognitive Neuroscience 26~(1) (2014) 132--142.

\bibitem{kriegeskorte2013representational}
N.~Kriegeskorte, R.~A. Kievit, Representational geometry: integrating
  cognition, computation, and the brain, Trends in Cognitive Sciences 17~(8)
  (2013) 401--412.

\bibitem{mur2013human}
M.~Mur, M.~Meys, J.~Bodurka, R.~Goebel, P.~A. Bandettini, N.~Kriegeskorte,
  Human object-similarity judgments reflect and transcend the primate-it object
  representation, Frontiers in Psychology 4 (2013) 128.

\bibitem{carlson2013representational}
T.~Carlson, D.~A. Tovar, A.~Alink, N.~Kriegeskorte, Representational dynamics
  of object vision: the first 1000 ms, Journal of Vision 13~(10) (2013) 1--19.

\bibitem{dosovitskiy2014discriminative}
A.~Dosovitskiy, J.~T. Springenberg, M.~Riedmiller, T.~Brox, Discriminative
  unsupervised feature learning with convolutional neural networks, in:
  Advances in Neural Information Processing Systems, 2014, pp. 766--774.

\bibitem{jones2014computer}
N.~Jones, Computer science: The learning machines, Nature 505~(7482) (2014)
  146--148.

\bibitem{girshick2014rich}
R.~Girshick, J.~Donahue, T.~Darrell, J.~Malik, Rich feature hierarchies for
  accurate object detection and semantic segmentation, in: IEEE Conference on
  Computer Vision and Pattern Recognition, 2014, pp. 580--587.

\bibitem{alemi2013multifeatural}
A.~Alemi-Neissi, F.~B. Rosselli, D.~Zoccolan, Multifeatural shape processing in
  rats engaged in invariant visual object recognition, The Journal of
  Neuroscience 33~(14) (2013) 5939--5956.

\bibitem{szegedy2014going}
C.~Szegedy, W.~Liu, Y.~Jia, P.~Sermanet, S.~Reed, D.~Anguelov, D.~Erhan,
  V.~Vanhoucke, A.~Rabinovich, Going deeper with convolutions, arXiv:1409.4842.

\bibitem{le2013building}
Q.~V. Le, Building high-level features using large scale unsupervised learning,
  in: IEEE International Conference on Acoustics, Speech and Signal Processing,
  2013, pp. 8595--8598.

\bibitem{mohamed2011deep}
A.-r. Mohamed, T.~N. Sainath, G.~Dahl, B.~Ramabhadran, G.~E. Hinton, M.~A.
  Picheny, Deep belief networks using discriminative features for phone
  recognition, in: IEEE International Conference on Acoustics, Speech and
  Signal Processing, 2011, pp. 5060--5063.

\bibitem{donahue2013decaf}
J.~Donahue, Y.~Jia, O.~Vinyals, J.~Hoffman, N.~Zhang, E.~Tzeng, T.~Darrell,
  Decaf: A deep convolutional activation feature for generic visual
  recognition, arXiv:1310.1531.

\bibitem{vanrullen2001time}
R.~Vanrullen, S.~J. Thorpe, The time course of visual processing: from early
  perception to decision-making, Journal of Cognitive Neuroscience 13~(4)
  (2001) 454--461.

\bibitem{fabre2011characteristics}
M.~Fabre-Thorpe, The characteristics and limits of rapid visual
  categorization., Frontiers in psychology 2~(243) (2011) 1--12.

\bibitem{kirchner2006ultra}
H.~Kirchner, S.~J. Thorpe, Ultra-rapid object detection with saccadic eye
  movements: Visual processing speed revisited, Vision Research 46~(11) (2006)
  1762--1776.

\bibitem{mack2011timing}
M.~L. Mack, T.~J. Palmeri, The timing of visual object categorization,
  Frontiers in Psychology 2~(165) (2011) 1--8.

\bibitem{potter2014detecting}
M.~C. Potter, B.~Wyble, C.~E. Hagmann, E.~S. McCourt, Detecting meaning in rsvp
  at 13 ms per picture, Attention, Perception, \& Psychophysics 76~(2) (2014)
  270--279.

\bibitem{kreiman2007limits}
G.~Kreiman, T.~Serre, T.~Poggio, On the limits of feed-forward processing in
  visual object recognition, Journal of Vision 7~(9) (2007) 1041.

\bibitem{roelfsema2002figure}
P.~R. Roelfsema, V.~A. Lamme, H.~Spekreijse, H.~Bosch, Figure—ground
  segregation in a recurrent network architecture, Journal of Cognitive
  Neuroscience 14~(4) (2002) 525--537.

\bibitem{raudies2010neural}
F.~Raudies, H.~Neumann, A neural model of the temporal dynamics of
  figure--ground segregation in motion perception, Neural Networks 23~(2)
  (2010) 160--176.

\bibitem{gilbert2013top}
C.~D. Gilbert, W.~Li, Top-down influences on visual processing, Nature Reviews
  Neuroscience 14~(5) (2013) 350--363.

\bibitem{pannunzi2012learning}
M.~Pannunzi, G.~Gigante, M.~Mattia, G.~Deco, S.~Fusi, P.~Del~Giudice, Learning
  selective top-down control enhances performance in a visual categorization
  task, Journal of Neurophysiology 108~(11) (2012) 3124--3137.

\bibitem{wyatte2012limits}
D.~Wyatte, T.~Curran, R.~O'Reilly, The limits of feedforward vision: Recurrent
  processing promotes robust object recognition when objects are degraded,
  Journal of Cognitive Neuroscience 24~(11) (2012) 2248--2261.

\bibitem{o2013recurrent}
R.~C. O'Reilly, D.~Wyatte, S.~Herd, B.~Mingus, D.~J. Jilk, Recurrent processing
  during object recognition, Frontiers in Psychology 4~(124) (2013) 1--14.

\bibitem{rajalingham2015comparison}
R.~Rajalingham, K.~Schmidt, J.~J. DiCarlo, Comparison of object recognition
  behavior in human and monkey, The Journal of Neuroscience 35~(35) (2015)
  12127--12136.

\bibitem{majaj2015simple}
N.~J. Majaj, H.~Hong, E.~A. Solomon, J.~J. DiCarlo, Simple learned weighted
  sums of inferior temporal neuronal firing rates accurately predict human core
  object recognition performance, The Journal of Neuroscience 35~(39) (2015)
  13402--13418.

\bibitem{sastyin2015does}
G.~Sastyin, R.~Niimi, K.~Yokosawa, Does object view influence the scene
  consistency effect?, Attention, Perception, and Psychophysics 77~(3) (2015)
  856--866.

\bibitem{oliva2007role}
A.~Oliva, A.~Torralba, The role of context in object recognition, Trends in
  Cognitive Sciences 11~(12) (2007) 520--527.

\bibitem{joubert2008early}
O.~R. Joubert, D.~Fize, G.~A. Rousselet, M.~Fabre-Thorpe, Early interference of
  context congruence on object processing in rapid visual categorization of
  natural scenes, Journal of Vision 8~(13) (2008) 1--11.

\bibitem{remy2013object}
F.~R{\'e}my, L.~Saint-Aubert, N.~Bacon-Mac{\'e}, N.~Vayssi{\`e}re, E.~Barbeau,
  M.~Fabre-Thorpe, Object recognition in congruent and incongruent natural
  scenes: a life-span study, Vision Research 91 (2013) 36--44.

\bibitem{isik2014dynamics}
L.~Isik, E.~M. Meyers, J.~Z. Leibo, T.~Poggio, The dynamics of invariant object
  recognition in the human visual system, Journal of Neurophysiology 111~(1)
  (2014) 91--102.

\bibitem{nishimura2014size}
M.~Nishimura, K.~Scherf, V.~Zachariou, M.~Tarr, M.~Behrmann, Size precedes
  view: developmental emergence of invariant object representations in lateral
  occipital complex., Journal of Cognitive Neuroscience 27~(3) (2015) 474--491.

\bibitem{farzmahdi2015specialized}
A.~Farzmahdi, K.~Rajaei, M.~Ghodrati, R.~Ebrahimpour, S.-M. Khaligh-Razavi, A
  specialized face-processing network consistent with the representational
  geometry of monkey face patches, arXiv:1502.01241.

\bibitem{tsao2003faces}
D.~Y. Tsao, W.~A. Freiwald, T.~A. Knutsen, J.~B. Mandeville, R.~B. Tootell,
  Faces and objects in macaque cerebral cortex, Nature Neuroscience 6~(9)
  (2003) 989--995.

\bibitem{tsao2006cortical}
D.~Y. Tsao, W.~A. Freiwald, R.~B. Tootell, M.~S. Livingstone, A cortical region
  consisting entirely of face-selective cells, Science 311~(5761) (2006)
  670--674.

\end{thebibliography}

\end{document}